RESEARCH-ARTICLE

# Individualized Federated Learning for Traffic Prediction with Error-Driven Aggregation

# Individualized Federated Learning for Traffic Prediction with Error-Driven Aggregation

HANG CHEN, Computer and Information Sciences, University of Delaware, Newark, United States
COLLIN MEESE, Department of Civil and Environmental Engineering, University of Delaware, Newark, United States
MARK NEJAD, Department of Civil and Environmental Engineering, University of Delaware, Newark, United States
CHIEN-CHUNG SHEN, Computer and Information Sciences, University of Delaware, Newark, United States

Low-latency traffic prediction is vital for smart city traffic management. Federated Learning has emerged as a promising technique for Traffic Prediction (FLTP), offering several advantages such as privacy preservation, reduced communication overhead, improved prediction accuracy, and enhanced adaptability to changing traffic conditions. However, the majority of the current FLTP frameworks lack a real-time model updating scheme, which hinders their ability to continuously incorporate new incoming traffic data and adapt effectively to the changing dynamics of traffic trends. Another concern with the existing FLTP frameworks is their reliance on the conventional FL model aggregation method, which involves assigning an identical model (i.e., the global model) to all traffic monitoring devices to predict their individual local traffic trends, thereby neglecting the non-IID characteristics of traffic data collected in different locations. Building upon these findings, we propose *NeighborFL*, an individualized real-time federated learning scheme that introduces a haversine distance-based and error-driven grouping heuristic for personalized local models from the perspective of each individual traffic node. This approach allows *NeighborFL to create location-aware and tailored prediction models for each client while fostering collaborative learning*. Simulations demonstrate the effectiveness of *NeighborFL*, offering improved real-time prediction accuracy over three baseline models, with one experimental setting showing a 16.9% reduction in MSE value compared to a naive FL setting.

## 1 Introduction

In the pursuit of high fidelity Traffic Prediction (TP), Federated Learning (FL) has emerged as a technique with promising performance in traffic flow prediction [31] and speed prediction [34]. Compared to the conventional centralized approach [16], FL presents a myriad of advantages that contribute to its growing popularity and potential impacts on TP. In particular, the distributed nature of FL allows for continuous model updates based on the latest traffic data collected at each traffic monitoring device, or device for short. This dynamic learning capability enables the model to promptly incorporate changes and adapt to evolving traffic patterns in real time,

Authors' Contact Information: Hang Chen, Computer and Information Sciences, University of Delaware, Newark, Delaware, United States; e-mail: chenhang@udel.edu; Collin Meese, Department of Civil and Environmental Engineering, University of Delaware, Newark, Delaware, United States; e-mail: cmeese@udel.edu; Mark Nejad, Department of Civil and Environmental Engineering, University of Delaware, Newark, Delaware, United States; e-mail: nejad@udel.edu; Chien-Chung Shen, Computer and Information Sciences, University of Delaware, Newark, Delaware, United States; e-mail: cshen@udel.edu.

ensuring that the predictions remain accurate and up to date. Such adaptability is essential to traffic prediction, as the dynamic nature of traffic patterns necessitates agile learning and rapid adjustments of the model [13]. In addition, FL has demonstrated its potential to enhance prediction accuracy, where the distributed nature of FL incorporates diverse data sources, and encompasses various traffic scenarios and conditions. This diversity contributes to a more comprehensive and robust training process, resulting in improved prediction accuracy. The collaborative aspect of FL also pools the knowledge from multiple traffic management organizations, creating collective intelligence that surpasses the capabilities of centralized models [20].

Early **F**ederated **L**earning frameworks for **T**raffic **P**rediction (FLTP) often adopted a naive aggregation approach. These frameworks aimed to produce a global model that would be utilized by all the devices or organizations within the study region to predict individual short-term traffic situations [14, 20, 24, 34]. However, this aggregation method overlooked one important factor: the traffic data collected across different devices could exhibit non independent and identically distributed (non-IID) data characteristics [26]. For instance, devices located in proximity to each other along the same arterial corridor would more likely exhibit similar traffic trends compared to those located farther apart, or on ramps or other roads, as observed in [20]. In other words, local models trained by respective devices may exhibit a significant bias towards their localized traffic trends [11]. This data imbalance limits the effectiveness of training a single, optimal global model, as it will likely not perform optimally for all devices.

To address the non-IID issue in FLTP, a common approach is to incorporate similarity clustering during the aggregation process [18]. Indeed, even non-FL-based TP frameworks have explored clustering methods, such as $K$-nearest-neighbors-based clustering, to enhance prediction accuracy [3, 7, 15], and more recent FL works have explored the $K$-means algorithm and Graph Neural Networks (GNN) based approaches for clustering the devices [14, 23, 35]. More specifically, the devices within the study region are divided into non-overlapping subgroups using specific metrics, such as geographical distance [3], prediction error [14], or model parameter similarity [35]. Although these approaches have been shown to enhance the accuracy of the global model employed by individual devices, a mere static grouping of devices may prove ineffective, particularly during unexpected and disruptive traffic events, such as accidents or disabled vehicles. These random, non-recurring events introduce a dynamic facet characterized by location-dependency and high-intensity, yet ephemeral impacts on the traffic network, prompting the requirement for an adaptive approach.

Another notable observation of many existing FLTP frameworks is that these federated models are typically trained in an offline manner [14, 23, 24, 32, 34, 35]. This implies that the federated models are exclusively trained using historical traffic data that has already been collected by the devices, thus relying solely on past information for learning purposes. Once the federation is completed, the final global model is deployed in the live environment to make predictions based on real-time streaming data. However, a practical TP model should adapt in real-time using newly incoming traffic data. This is crucial to effectively handle data drift caused by the changing traffic dynamics, such as non-recurring events, seasonal cycles, and infrastructure changes. It is worth noting that FL has demonstrated the ability to update models in real-time, as exemplified by its successful implementation in the renowned Gboard word prediction application [30]. Therefore, it is essential to explore the feasibility of FLTP in a real-time streaming manner.

Motivated by the aforementioned insights and drawing inspiration from related research [17], we propose ***NeighborFL***, a real-time and individualized FL-based TP workflow to further improve the prediction accuracy of each individual traffic sensing location. This workflow introduces a new individualized aggregation method, guided by minimizing real-time prediction error and utilizing distances between devices as a heuristic. In ***NeighborFL***, instead of having one global model for all devices in each communication round, each device uses an individualized group of local models from its chosen neighboring devices, plus its own local model, to create its aggregated model for real-time prediction.

From a high level perspective, *NeighborFL* operates as follows:

(1) During the initialization phase of the federation, each device begins with an initial model and an empty set to record its chosen *Favorite Neighbors* within a predefined radius. The initial model can be randomly initialized or pretrained using the device's historical traffic data. All devices adhere to the same model architecture, following the convention in the original FL framework [19]
(2) At the start of each communication round, every device has an individualized model, denoted as $A$, which is produced by aggregating the parameters of the local models obtained from the devices in its *Favorite Neighbors* set (short for *FN*) along with its own local model (denoted as $L_{own}$) from the previous communication round. For the first communication round, each device's *FN* is empty, and the initial model serves as the individualized model. In contrast to conventional FL approaches, where devices share the same global model, denoted as $G$, for traffic prediction, each device employs its $A$ to make predictions, labeled as $P$, while simultaneously gathering new traffic data from its sensors in real-time.
(3) In addition, a device might have another aggregated model, denoted as $A_{eval}$, in situations where its *FN* has not included all its candidate neighbors and the device decides to evaluate a neighboring device's model for potential inclusion as a *Favorite Neighbor* in the current round. This is done by acquiring the local model from its nearest device not yet included in its *FN* at the end of the last communication round. This specific local model is referred to as $L_{eval}$, and the owner of $L_{eval}$ is denoted as $d_{eval}$. The production of $A_{eval}$ involves aggregating the parameters of the local models gathered from its *Favorite Neighbors*, and those of $L_{own}$ and $L_{eval}$. In essence, the difference between $A_{eval}$ and $A$ is the inclusion of an extra local model (i.e., $L_{eval}$) in the aggregation process of $A_{eval}$.
(4) The process of evaluating $L_{eval}$ involves generating a new set of predictions with $A_{eval}$, which we refer to as $P_{eval}$. The device then assesses the accuracy of these predictions by comparing the prediction errors between $P$ and $P_{eval}$, using an error metric such as the mean squared error. This comparison involves evaluating the error of $P$ against the actual ground truth, $T$, which we denote as $E$, and similarly, evaluating the error of $P_{eval}$ against $T$, denoted as $E_{eval}$.
(5) If $E_{eval}$ is smaller than $E$, indicating that the inclusion of $L_{eval}$ in the model aggregation results in improved prediction accuracy in this round, $d_{eval}$ is included in the device's *FN*. Subsequently, $A_{eval}$ is adopted for local learning in the current round. Otherwise, $d_{eval}$ is skipped from being added to the device's *FN*, and is optionally placed in a retry list with an assigned retry interval, while $A$ is used for learning to produce the device's local model for the current round.
(6) To further adapt to real-time traffic dynamics, a device retains the option of eliminating one or more devices from its *FN* based on certain conditions. For instance, this might occur when a device's *FN* becomes stable (i.e., not changing for a series of continuous rounds), or when the real-time prediction error consistently increases over several consecutive rounds.
(7) Each device repeats Steps 2-6 in each subsequent communication round, adapting to the evolving traffic dynamics and refining its prediction capabilities in a dynamic and collaborative manner.

Figure 1 depicts the above steps of *NeighborFL* throughout a communication round, utilizing mean squared error as the error metric and adopting *FedAvg* [19] for local model aggregation. Additionally, it offers a parallel comparison with the workflow of the conventional FL approaches.

The objective of *NeighborFL* is to enable devices to collaboratively search for an improved model that leads to lower prediction errors in the present round. This collaborative effort takes into account the spatial relationship among devices, aiming to enhance the predictions made in the subsequent round considering the temporal conditions. The contributions of *NeighborFL* and the distinct differences from existing FLTP architectures are summarized as follows:

- ***Personalized Aggregated Models.*** Instead of starting the learning process from a single global model generated in the previous communication round, our approach first employs a personalized aggregated model for prediction and another aggregated model for evaluation if a $A_{eval}$ is proposed. Then, each device tests these two models and selects the one with the lower prediction error to continue learning. This approach enables the system to adapt more effectively to real-time traffic dynamics.

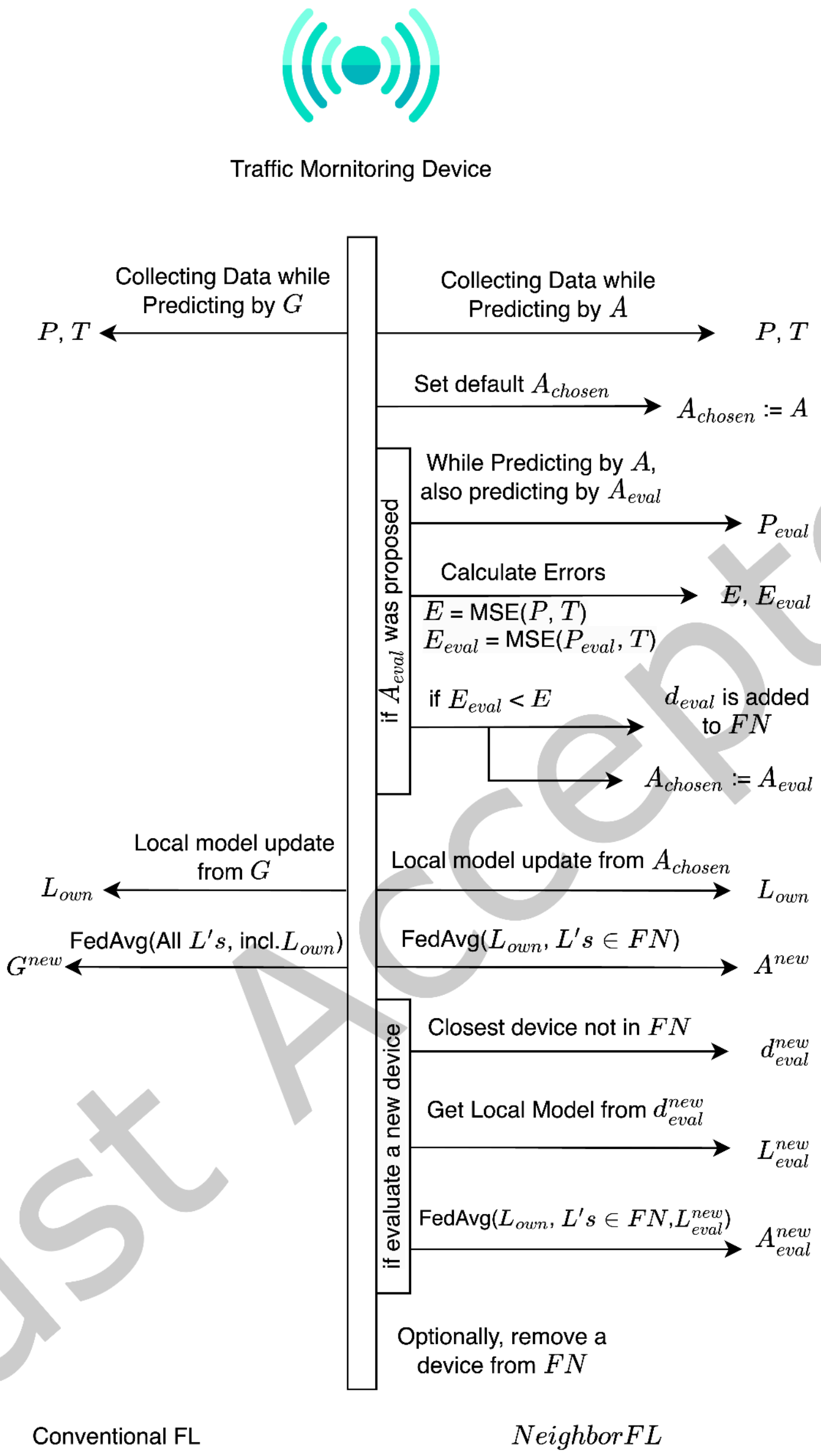

Fig. 1. A high-level overview of *NeighborFL* and its comparison with the Conventional FL approach.

- ***Radius-based Candidate Selection.*** A device can set a radius to confine its candidate *Favorite Neighbors*, limiting the maximum number of local models that will be used to produce the aggregated model. This restriction prevents a device from requesting the local models of spatially distant devices, thereby avoiding potential communication delays, exploiting cooperative edge computing infrastructure, and improving the response time of the entire system.
- ***Knowledge Propagation.*** The *Favorite Neighbors* sets of two nearby devices may contain overlapping devices. As a result, a local model from one device can still propagate its learned knowledge to a distant device, even if the distant device does not have the corresponding owner in its *Favorite Neighbors* set. This propagation of knowledge across devices employs natural weightings to local models during aggregation, facilitating the dissemination of valuable insights and, thereby, enhancing the overall predictive capabilities of the system.
- ***Individual Device Perspective.*** *NeighborFL* allows the devices to explore the spatial and temporal correlation between other participants to form dynamic and individualized groups, rather than considering the entire cohort of devices statically. This device-centric approach builds upon prior research and has the potential to directly incorporate decentralized, device-level model cross-validation to enhance system robustness and resilience [5].

We conducted comprehensive simulations to evaluate the effectiveness of *NeighborFL*. Without the loss of generality, we chose traffic speed data from 26 detectors in the publicly available PEMS-BAY reference dataset [12], and used the Long Short-Term Memory (LSTM) [10] recurrent neural network as the client model architecture. In addition to using a randomly initialized global model across each of the 26 devices, we also investigated the impact of pretraining the initial model in *NeighborFL* [27]. For this, we allowed each of the 26 devices to pretrain their initial models using the traffic speed data from the first week of January 2017. We then analyzed the real-time learning and prediction capabilities of *NeighborFL* by testing its performance over the following two-week period involving both the pretrained and non-pretrained initial models. The experimental results presented in Section 4 demonstrate that *NeighborFL* outperforms the three baseline models in terms of real-time prediction capability in the majority of cases, as indicated by the lower prediction error. Our *NeighborFL* code repository is available at https://github.com/hanglearning/NeighborFL , and our experiments can be reproduced by the provided portion of the dataset and the specified random seed.

The remainder of this paper is organized as follows. Section 2 provides a comprehensive review of related work that has guided us in the development of *NeighborFL*. Section 3 presents the problem definitions for the key components of *NeighborFL* and provides a detailed outline of the workflow. Experimental results are presented in Section 4. Section 5 addresses certain limitations of the current *NeighborFL* design and proposes potential directions for future research. Section 6 summarizes the key findings of the paper.

## 2 Related Work

In this section, we review the existing TP architectures by categorizing them into three general categories - centralized methods, non-streaming FL methods, and streaming FL methods. Both centralized and non-streaming FL methods are further differentiated into methods with and without the grouping mechanism.

### 2.1 Centralized methods

*2.1.1 Non-grouping.* Research in TP has a long history and has become a crucial component of Intelligent Transportation Systems. Early conventional TP models, such as the autoregressive integrated moving average (ARIMA) [1] and exponential smoothing [28], have been used to forecast short-term traffic based on previous observations. In recent years, deep learning-based approaches have gained popularity. Lv et al. (2015) introduced a stacked autoencoder model [16], while Fu et al. (2016) were among the first to utilize LSTM and GRU neural networks [8] for traffic flow prediction. The LSTM and GRU models proposed in [8] demonstrated superior prediction performance compared to the ARIMA model, solidifying the foundation of recurrent neural networks in TP.

*2.1.2 Grouping.* The $K$-nearest neighbor (KNN) is one of the most widely used grouping mechanisms in the TP literature. One of the first uses of KNN in TP research can be found in [7], where its capability for improving TP was analyzed using freeway data. The forecasting method with KNN involves computing the average of predictions from a traffic node's $K$-nearest neighbors and its own. However, the authors concluded that "the KNN method performed comparably to, but not better than, the linear time-series approach, and further research is needed to delineate those situations where the KNN approach may be preferable." In 2012, Chang et al. proposed a KNN-based non-parametric regression (KNN-NPR) TP model to address situations where current or future traffic data exhibits fluctuations or abrupt changes. The proposed model demonstrated effectiveness and easy optimization for minimizing prediction errors [3]. Furthermore, Luo et al. (2019) introduced a model named KNN-LSTM, which utilized KNN to select the most relevant neighboring traffic nodes with respect to a specific node. The final prediction results are obtained by weighting the LSTM prediction values from all $K$ stations in addition to its own output[15]. We observed that *NeighborFL* shares a similar methodology with KNN-LSTM in terms of fusing the prediction values. However, in KNN-LSTM, the $K$ value specifying the number of neighboring traffic nodes to consider for any given node remains constant. In contrast, our *Favorite Neighbors* set dynamically expands and shrinks as communication rounds progress.

These works have demonstrated that considering the spatiotemporal correlation characteristics of traffic data can improve the prediction performance of individual nodes. They also inspired us to incorporate distance as a heuristic when a traffic node searches for a *Favorite Neighbor*.

## 2.2 Non-streaming FL methods

*2.2.1 Non-grouping.* An early application of FLTP was introduced by Liu et al. (2020) [14], where the authors proposed a FL-based architecture for TP using Gated Recurrent Units (GRU) [6], termed FedGRU. FedGRU demonstrated comparable prediction performance to centralized GRU methods while addressing privacy concerns associated with the traffic dataset. Qi et al. (2021) further advanced FLTP by integrating it into a blockchain architecture [24]. This framework addresses the single-point-of-failure issue in centralized FLTP and leverages smart contract technology to safeguard against malicious attacks and ensure high-quality training data. Zhang et al. (2021) extended FLTP by incorporating Graph Neural Networks (GNN) and proposing an adjacency matrix aggregation approach, enabling local GNN-based models to access the global network for improved training effectiveness [34].

*2.2.2 Grouping.* The same work by Liu et al. (2020) introduced an ensemble clustering-based FedGRU algorithm that utilizes traffic data with improved spatio-temporal correlation based on location information. This approach enhances prediction accuracy compared to the standard FedGRU method. It also handles scenarios where multiple clients collaborate to train a traffic prediction model by grouping traffic nodes into $K$ clusters prior to applying FedGRU [14]. Zeng et al. (2021) developed a divisive hierarchical clustering approach to partition traffic data at each traffic node into clusters. They applied FL to collaboratively train learning models for each cluster across all stations. The authors demonstrated improved prediction performance compared to the standard FL method using the PeMS dataset [32]. Furthermore, GNN-based methods are also explored as heuristics to group traffic nodes into non-overlapping clusters in [35] and [23].

Although these FLTP frameworks paved the way for exploring privacy-preserving distributed learning in TP, they lack the ability to update the global model in real-time after deployment, making them unable to adapt to changing traffic dynamics, such as non-recurrent traffic incidents and varying traffic patterns. We categorize these FLTP frameworks as non-streaming FL methods since they do not use real-time streaming traffic data for training or updating the models.

## 2.3 Streaming FL methods

In contrast to non-streaming FL approaches, streaming FL architectures rely on real-time traffic data streams to provide inputs for learning and evaluation, allowing for live updates and continuous adaptation. For traffic prediction, this streaming setting is particularly challenging because traffic data exhibit strong spatial–temporal correlations, abrupt regime shifts due to non-recurrent events (e.g., incidents or special events), and periodic patterns driven by daily and weekly demand cycles. These properties create non-stationary, highly heterogeneous data streams across sensors that amplify the usual FL challenges of concept drift, client imbalance, and non-IID data.

Our research identified two streaming FLTP models. Meese et al. (2022, 2024) introduced BFRT, a streaming Blockchained FL workflow for TP that defines a protocol for federated models to predict real-time traffic volume while collecting live data. By comparing it with a per-device centralized model, the authors demonstrated that the streaming FL method exhibits lower real-time prediction error, highlighting the advantages of collaborative real-time learning [20, 21]. Subsequently, Liu et al. (2023) proposed FedOSTC, an alternative streaming FLTP workflow that incorporates a Graph Attention Network to assess spatial correlation among traffic nodes. It also integrates a period-aware aggregation mechanism to combine local models optimized using the Online Gradient Descent (OGD) algorithm [13]. FedOSTC outperforms five chosen baselines in terms of prediction performance on the partial PEMS-BAY and METR-LA datasets [29].

To the best of our knowledge, there has been no published work focusing on streaming FLTP methods involving grouping, i.e., personalized aggregation, and *NeigborFL* aims to fill this research gap and lays the foundation for multiple research directions discussed in Section 5.

# 3 System Modules

In this section, we present *NeighborFL's* key modules and their formulations, along with their implementation in pseudocode. The key mathematical notations used in this section are summarized in Table 6 in the Appendix.

## 3.1 System Entity

*NeighborFL* is operated by a set of traffic monitoring devices $\mathcal{D} = \{d_1, d_2, \cdots, d_i, \cdots\}$ in the study region through communication rounds $\mathcal{R} = \{R_1, R_2, \cdots, R_j, \cdots\}$. Each device $d$ has a unique device ID and sequence number. We assume that a device is equipped with sensor(s) to collect traffic data such as volume, speed, and occupancy, and we also assume a device can act as a computational node to perform model training and participate in other algorithmic operations at the network edge.

## 3.2 Candidate and Favorite Neighbors

The concept of *Favorite Neighbors* distinguishes *NeighborFL* from other FLTP frameworks. In *NeighborFL*, a device aggregates local models from itself plus its *Favorite Neighbors* (Algo. 10: line 21) instead of using the local models from all the participants (i.e., $\mathcal{D}$). Prior to joining *NeighborFL*, a device initializes an empty set called *FN* to keep track of its **favorite neighbors** (Algo. 10: line 5) and a radius value $r$ to populate a hashmap called *CFN* to record its **candidate favorite neighbors** (Algo. 1: lines 12, 17).

A single candidate favorite neighbor is denoted as $cfn$. A device considers neighboring devices within a distance of $r$ as candidates. The key in *CFN* represents the device ID of the $cfn$, while the associated value is either (i) the Haversine distance between the $cfn$ and the device (Algo. 1: lines 9, 12), or (ii) a *bucket index* indicating proximity when the GPS coordinates of devices need to be kept private (Algo. 1: lines 14, 17). In the latter case, a smaller bucket index means closer, and no raw GPS coordinates or exact distances are revealed or stored. In this work, a device's $cfn$ becomes its favorite neighbor ($fn$) after evaluating the real-time prediction error, as described in Sec. 3.3.4. The evaluation order is either determined by the Haversine distance between the device and $cfn$ **from closest to farthest**, or by the bucket index **from lowest to highest** (Algo. 1: line 21; Algo. 2: line 4), with a retry interval control mechanism (Algo. 1: lines 4-6, 18-20; Algo. 2: lines 5-8; Algo. 6: lines 15-17) explained in Sec. 3.3.4. Algo. 1 outlines the process of populating *CFN* for the $i$-th device $d_i$, while Algo. 2 explains how a $cfn$ may be selected by $d_i$ in $R_j$ for evaluation based on the relative Haversine distance or bucket index. Note that different devices may choose different values for $r$, and $r_i$ specifically represents the $r$ value chosen by device $d_i$ in Algo. 1. $d^j_{i,eval}$ represents the selected $cfn$ to be evaluated by $d_i$ in $R_{j+1}$ (Algo. 2: line 7). Fig. 2 illustrates that within the radius $r_i$ around $d_i$, the candidate favorite neighbors are $d_1$–$d_4$. Devices $d_5$–$d_7$ lie outside $r_i$ and are therefore excluded from $d_i$'s *CFN*; their models are not evaluated in downstream processing.

---

**Algorithm 1:** FormCFN() - $d_i$ forms candidate favorite neighbors hash map *CFN*

1 **Input:** $r_i$
2 **Output:** None
3 $d_i.CFN \leftarrow \{\}$;
  `/* An ordered hash map. */`
  `/* Key: A` $cfn$`'s device ID */`
  `/* Value: Haversine distance between` $d_i$ `and the` $cfn$ `*/`
4 $d_i$.last_try_round ← {};
5 $d_i$.retry_interval ← {};
6 $d_i$.rep_book ← {};
7 **for** *each* $d \in \mathcal{D} \setminus d_i$ **do**
8     **if** *d.GPS is available* **then**
9         distance = HaversineDistance($d$.GPS, $d_i$.GPS);
10         **if** *distance* $> r_i$ **then**
11             continue to the next $d$;
12         $d_i.CFN[d]$ = distance;
13     **else**
        `/* Algo. 1a */`
14         $j \leftarrow$ BucketIndexGC($d_i, d, r_i$);
15         **if** $j = \perp$ **then**
16             continue to the next $d$;
17         $d_i.CFN[d] = j$;
18     $d_i$last_try_round[$d$] ← 0;
19     $d_i$.retry_interval[$d$] ← 0;
20     $d_i$.rep_book[$d$] ← 0;
21 Sort($d_i.CFN$)
  `/* Sort` $d_i$`'s` $cfn$`'s by Harversine distance from close to far, or by bucket index from low to high if raw GPS is private. */`

---

In situations where the GPS coordinates of devices need to be kept private, we replace raw GPS coordinates and exact Haversine distances with a two-party, privacy-preserving within-radius predicate over a small set of radii "buckets". Concretely, to decide whether $d$ is a $cfn$ for $d_i$, we evaluate—over increasing radii $\beta_1 < \beta_2 < \cdots < \beta_K = r_i$—the predicate

$$(\Delta\text{lon} \cdot \cos(\text{lat}_i))^2 + (\Delta\text{lat})^2 \leq (\beta/69.093)^2$$

inside a Yao garbled-circuit protocol [25], which reveals only the one-bit outcome (Algo. 1a: line 11) and discloses neither party's GPS coordinates. Here $\Delta\text{lat} = \text{lat}_d - \text{lat}_i$ and $\Delta\text{lon} = \text{lon}_d - \text{lon}_i$ are in degrees, and 69.093 is the approximate number of miles per degree of latitude (north–south); the east–west scale difference is handled by $\cos(\text{lat}_i)$, computed within the circuit. Device $d_i$ acts as the generator, garbling the circuit and sending garbled tables and its input labels (Algo. 1a: lines 7-9); device $d$ acts as the evaluator, obtains its input labels via 1-out-of-2 Oblivious Transfer, evaluates, and returns only the output label for $d_i$ to decode as a single bit (inside radius or not) (Algo. 1a: lines 9-11). We record in *CFN* the index of the smallest passing bucket (Algo. 1a: lines 12-13; Algo. 1: lines 14, 17) and order candidates by bucket index from lowest to highest. The retry-interval logic and downstream selection remain unchanged.

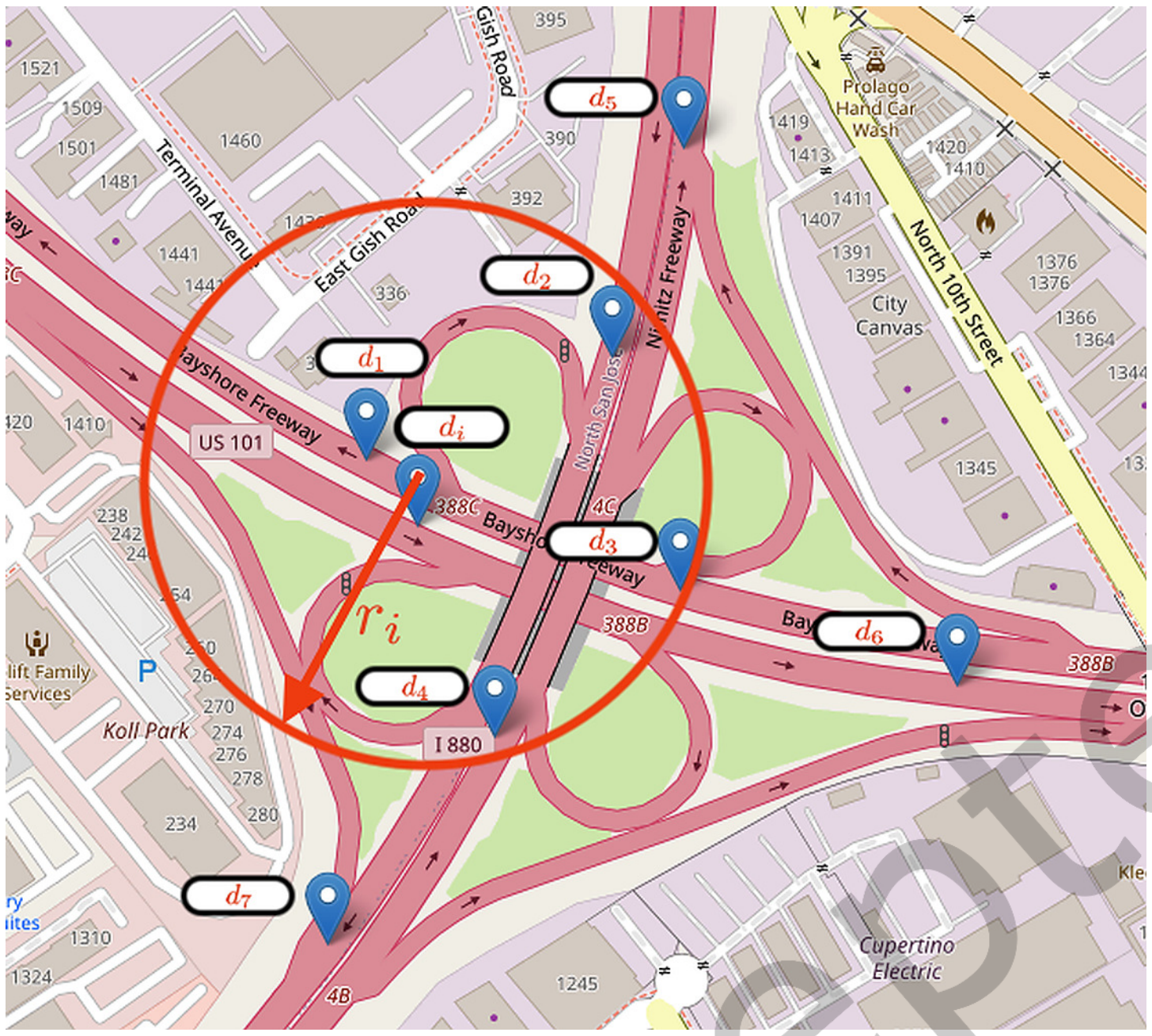

Fig. 2. Candidate favorite neighbors of $d_i$ within $r_i$: $d_1$, $d_2$, $d_3$ and $d_4$.

Technically, a device has the option to set $r$ as unlimited to consider all devices in the study region. However, selecting an appropriate value for $r$ can enhance a device's operational efficiency, and the overall system scalability, by restricting the number of candidate neighbors. This is important because distant devices located on different roads may display distinctly different traffic patterns, decreasing the evaluation efficiency. Also note that a device might consider evaluating more than one candidate. Without loss of generality, we concentrate on the scenario where only one $cfn$ is evaluated at a time in this study (Algo. 2, lines 6-8).

**Algorithm 1a:** BucketIndexGC($d_i, d, r_i$) — smallest passing bucket via Two-Party Computation

**Input:** $d_i$'s ($\text{lat}_i, \text{lon}_i$); $d$'s ($\text{lat}_d, \text{lon}_d$); $r_i$
**Output:** $j^\star \in \{1, \ldots, K\}$ or $\perp$
/* $j^\star$ is the smallest passing bucket index; $\perp$ means no bucket passed. */
*Notation:* $\Delta\text{lat} = \text{lat}_d - \text{lat}_i$, $\Delta\text{lon} = \text{lon}_d - \text{lon}_i$ (degrees); MPD = 69.093 miles/degree
/* MPD = Miles Per Degree (latitude); east-west scale uses cos($\text{lat}_i$). */
$\mathcal{B}_i \leftarrow$ MakeBuckets($r_i$);
$K \leftarrow |\mathcal{B}_i|$ /* Number of buckets. */
**for** $j \leftarrow 1$ **to** $K$ **do**
  gid $\leftarrow d_i$.GC_GarbleSend($d$, $C_{s \le t}(\beta_j)$);
  /* Generator $d_i$ garbles circuit for radius $\beta_j$ and sends to evaluator $d$. gid is the per-invocation garbled-circuit session identifier and $C_{s \le t}(\beta)$ is the in-circuit predicate. */
  $d_i$.GC_LockOutputToGenerator(gid);
  /* Lock output so only generator $d_i$ can decode the final bit. */
  $d_i$.OT_SendLabels(gid); $d$.OT_RecvLabels(gid);
  /* Oblivious Transfer: $d_i$ sends label pairs; $d$ receives one label per input bit. */
  $\ell_{\text{out}} \leftarrow d$.GC_Evaluate(gid);
  /* Evaluator $d$ runs the garbled circuit and obtains an opaque output label. */
  $b \leftarrow d_i$.GC_DecodeBit(gid, $\ell_{\text{out}}$);
  /* Generator $d_i$ decodes the returned label into the 1-bit result. */
  **if** $b == 1$ **then**
    **return** $j$
**return** $\perp$
/* $C_{s \le t}(\beta)$ implements $b \leftarrow \mathbf{1}\left[(\Delta\text{lon} \cdot \cos(\text{lat}_i))^2 + (\Delta\text{lat})^2 \le (\beta/\text{MPD})^2\right]$ in fixed-point (Q32.32). */

**Algorithm 2:** SelectCandidate() - $d_i$ in $R_j$ selects $cfn$ from $d_i$.*CFN* for evaluation

**Input:** None;
**Output:** $d^j_{i,eval}$ or None;
**if** $d_i$.FN.*size* < $d_i$.CFN.*size* **then**
  **for** *each* $cfn$ *in* $d_i$.CFN \ $d_i$.FN **do**
    /* by order of Harversine distance from close to far, or bucket index from low to high */
    $l = d_i$.last_try_round[$cfn$];
    **if** $l + d_i$.*retry_interval*[$cfn$] < $j$ **then**
      $d^j_{i,eval} \leftarrow cfn$;
      exit;

## 3.3 Online Streaming Model

*NeighborFL* is designed to operate in real-time regarding data collection, training, and prediction. This section provides formal definitions for these three aspects.

*3.3.1 Online Streaming Data Collection.* Each device collects traffic data in real-time for training, prediction, and evaluation when participating in *NeighborFL*. We denote $x$ as a single traffic data point, and $x_{i,m}^{j}$ as the $m$-th traffic data point collected by device $d_i$ in communication round $R_j$. This data point, $x_{i,m}^{j}$, represents a scalar value such as traffic volume or speed (Algo. 3: line 6). We define $\tau^j$ as the total number of data points that a device needs to collect within the $R_j$ (Algo. 3: line 5). Each device is equipped with a memory card to store the collected data, and the collection of the data points stored in the memory of $d_i$ is represented as $data_i$. To mitigate overfitting with outdated data, the length of $data_i$ is limited to a maximum size called *MaxDataSize*. This ensures that data from older rounds is excluded from training in the current round (Algo. 3: lines 9-11). The algorithm for real-time data collection and updating $data_i$ is presented in Algo. 3. $data_i^j$ represents the updated $data_i$ during or after the completion of $R_j$.

**Algorithm 3:** UPDATEDATASET() - Real-time traffic data collection and update of $d_i$ in $R_j$

1 **Input:** $data_i^{j-1}$, $\tau^j$, *MaxDataSize*
2 **Output:** $data_i^j$
3 $m \leftarrow 1$;
4 $data_i^j \leftarrow data_i^{j-1}$;
5 **while** $m \le \tau^j$ **do**
6     $x_{i,m}^{j} \leftarrow d_i$.COLLECTDATA();
7     $data_i^j \leftarrow data_i^j$.APPEND($x_{i,m}^{j}$);
8     $m \leftarrow m + 1$;
9 **if** $data_i^j$.*size* > *MaxDataSize* **then**
10     *RemoveSize* ← $data_i^j$.*size* − *MaxDataSize*;
11     $data_i^j$.REMOVEOLDDATA(*REMOVESIZE*);

*3.3.2 Online Training.* As mentioned in Sec. 1, at the end of a communication round, a device utilizes the selected aggregated model and the updated $data$ for training and generating its updated local model (Algo. 4, lines 7-11). We define a training instance as $I = \{\mathbf{X}, y\}$, where $\mathbf{X} \subset data$ represents a vector containing a continuous sequence of traffic data points from $data$. The variable $y$ can be either a scalar value or a vector, depending on the desired number of output forecasting horizon (i.e., prediction steps) $\mathcal{O}$, and the length of $\mathbf{X}$ depends on the number of input units $\mathcal{I}$ of the model. At any moment, the number of training instances in $data$ can be calculated as

$$num(I) = data.size - \mathcal{I} - \mathcal{O} + 1 \quad (1)$$

For example, in this work, we employed an LSTM neural network with 12 input units ($\mathcal{I}$ = 12) and 1 output unit ($\mathcal{O}$ = 1), representing a 1-step look-ahead prediction. Let $x^{(m)}$ be the $m$-th data point in $data$ and $I^{(k)}$ be the $k$-th training instance. Suppose a device has collected at least *MaxDataSize* data points and consider *MaxDataSize* = 72, then the first training instance in $data$ can be expressed as $I^{(1)} = \{< x^{(1)}, x^{(2)}, ..., x^{(12)} >, x^{(13)}\}$, and the last training instance, corresponds to the $72 - 12 - 1 + 1 = 60$-th instance as calculated in Eq. 1, can be written as $I^{(60)} = \{< x^{(60)}, x^{(61)}, ..., x^{(71)} >, x^{(72)}\}$. The process of updating the local model of $d_i$ in $R_j$ using the defined local epoch numbers $\mathcal{E}$ and a batch size of 1 is outlined in Algo. 4.

**Algorithm 4:** LocalTraining() - $d_i$ in $R_j$ trains the selected aggregated model and produces local model

1 **Input:** Selected aggregated model $A_i^j$, $data_i^j$, $\mathcal{E}$, $\mathcal{I}$, $\mathcal{O}$
2 **Output:** Local model $L_i^j$
3 $L_i^j \leftarrow A_i^j$;
4 $k \leftarrow 1$;
5 $N \leftarrow num(\mathcal{I})$ as calculated by equation (1);
6 **if** $data_i^j.size \geq \mathcal{I} + \mathcal{O}$ **then**
7     **for** *total* $\mathcal{E}$ *epochs* **do**
8         **while** $k \leq N$ **do**
9             $I^{(k)} \leftarrow$ ExtractInstance($data_i^j$, $k$);
10             $L_i^j \leftarrow$ ModelTraining($L_i^j$, $I^{(k)}$);
11             $k \leftarrow k + 1$;

*3.3.3 Real-time Prediction.* In round $R_j$, device $d_i$ makes one real-time prediction for the next $\mathcal{O}$ time steps (denoted as $\hat{y}$) using an aggregated model from $R_{j-1}$, denoted as $A_i^{j-1}$, on the most recent $\mathcal{I}$ data points from $data_i^j$ (denoted as $\tilde{\mathbf{X}}$) (Algo. 5: lines 10-12), prior to collecting a new data point (Algo. 5: lines 13-14). Consequently, the number of predictions made by a device within **the second communication round ($R_2$) and onwards** is equal to the number of newly collected data points, which is denoted by $\tau^{2+}$ (Algo. 3: line 5; Algo. 5: lines 6-14). As a special case, in the first communication round ($R_1$), a device needs to collect at least $\mathcal{I}$ data points to make the first prediction (Algo. 5: line 7-8), and $\mathcal{I} + \mathcal{O}$ data points to initiate local learning[1] (Algo. 4, line 6, Algo. 5: line 15). Therefore, $\tau^1$ must be at least $\mathcal{I} + \mathcal{O}$ to ensure a local model is produced at the end of $R_1$. Note that $data_i^j.size$ denotes the number of data points (i.e., $x$'s) stored in $data_i^j$, and the maximum value of $data_i^j.size$ is *MaxDataSize*. Consequently, the number of predictions a device makes in $R_1$ would equal to

$$\tau^1 - \mathcal{I} - \mathcal{O} + 1 \tag{2}$$

which is similar to the calculation of the number of data instances in $data$ as expressed in Eq. 1.

The steps for real-time prediction by $d_i$ in $R_j$ are described in Algo. 5. The resulting ground truth $Y_i^j$ consists of instances of actual traffic data points (denoted as $y$) corresponding to the prediction instances (i.e., $\hat{y}$) in $P_i^j$. Specifically, each ground truth instance $y$ (Algo. 5: line 16) or prediction instance $\hat{y}$ (Algo. 5: line 11) is represented by a scalar value or a vector, depending on the value of $\mathcal{O}$, and is recorded in chronological order in $Y_i^j$ or $P_i^j$, respectively, controlled by $m$ over time within a communication round (Algo. 5: lines 6-7, 15, 18). $P_i^j$ and $Y_i^j$ will be utilized to demonstrate the evaluation of a candidate favorite neighbor in Sec. 3.3.4.

*3.3.4 Evaluation of a candidate neighbor.* As mentioned in Sec. 3.2, a device's candidate favorite neighbor ($cfn$) becomes its favorite neighbor ($fn$) based on real-time prediction error evaluation (Algo. 6: lines 11-12). Inspired by the concept of reinforcement learning and the general approach to training a neural network by minimizing inference errors, the following process occurs:

If a device $d_{i,eval}^j$ is selected for evaluation by device $d_i$ in round $R_j$ (Algo. 2: line 7), $d_i$ in round $R_{j+1}$ will evaluate an aggregated model $A_{i,eval}^j$ that integrates the local model of $d_{i,eval}^j$, denoted as $L_{i,eval}^j$ (Algo. 10: lines 18-19, 24-25). This evaluation involves the comparison of the prediction error $E^{j+1_{i,eval}}$, derived by applying an

[1]To ensure that there is a sufficient amount of data for training, it can be inferred that the value of *MaxDataSize* also needs to be larger than or equal to $\mathcal{I} + \mathcal{O}$.

**Algorithm 5:** PREDICT() - $d_i$ in $R_j$ predicts in real-time by the last selected aggregated model in $R_{j-1}$

**Input:** Aggregated model $A_i^{j-1}$, $data_i^{j-1}$, $\tau^j$, $\mathcal{I}$, $\mathcal{O}$
**Output:** Real-time Predictions $P_i^j$, Ground Truths $Y_i^j$
$m \leftarrow 1$;
$data_i^j \leftarrow data_i^{j-1}$;
$P_i^j \leftarrow []$; /* An empty ordered list */
$Y_i^j \leftarrow []$; /* An empty ordered list */
**while** $m \leq \tau^j$ **do**
  **if** $j == 1$ *and* $m \leq \mathcal{I}$ **then**
    continue to collect;
  **else**
    $\tilde{X} \leftarrow$ EXTRACTLATEST($data_i^j$, $\mathcal{I}$);
    $\hat{y} \leftarrow A_i^{j-1}$.INFERENCE($\tilde{X}$, $\mathcal{O}$);
    $P_i^j$.APPEND($\hat{y}$);
  $x_{i,m}^j \leftarrow d_i$.COLLECTDATA();
  $data_i^j \leftarrow data_i^j$.APPEND($x_{i,m}^j$);
  **if** $j > 1$ *or* ($j == 1$ *and* $m \geq \mathcal{I} + \mathcal{O}$) **then**
    $y \leftarrow$ EXTRACTLATEST($data_i^j$, $\mathcal{O}$);
    $Y_i^j$.APPEND($y$) ; // append the latest $\mathcal{O}$ data points as a ground truth instance from the real-time updated $data_i^j$
  $m \leftarrow m + 1$;

error metric between $P_{i,eval}^{j+1}$ and $Y_i^{j+1}$ (Algo. 6, line 7), and another prediction error $E_i^{j+1}$, obtained by assessing $P_i^{j+1}$ against $Y_i^{j+1}$ (Algo. 6, line 8). $P_{i,eval}^{j+1}$ is comprised of the predictions made by $A_{i,eval}^j$ (Algo. 10: lines 15-16), which includes $L_{i,eval}^j$ while being produced (Algo. 10: lines 24-25), whereas $P_i^{j+1}$ are composed of the predictions made by $A_i^j$ (Algo. 10: line 14), which does not include $L_{i,eval}^j$ (Algo. 10: line 21). In simpler terms, a device determines the effectiveness of a potential candidate neighbor device by incorporating its local model into an aggregated model for making predictions. It then compares the error from these predictions to the error from another aggregated model that excludes the neighbor's local model. The prediction error metric (Algo. 6, lines 7-10) used for evaluation could be any commonly used measure in traffic prediction models, such as MAE, MRE, MSE, RMSE, among others [14, 16]. An example of device-level MSE calculation for $\mathcal{O} = 1$ is given in Eq. 3.

If $E_{i,eval}^{j+1}$ is found to be lower than $E_i^{j+1}$, indicating that $L_{i,eval}^j$ contributes positively to the prediction in $R_{j+1}$, $d_i$ adds $d_{i,eval}^j$ to its favorite neighbor set $d_i.FN$ (Algo. 6: lines 11-12), and $A_{i,eval}^j$ will be chosen to perform local update at the end of $R_{j+1}$ (Algo. 6: line 13), meanwhile $L_{eval}^{j+1}$ will be included to produce $A_i^{j+1}$ (Algo. 10: line 25). Otherwise, $d_i$ records $d_{i,eval}^j$ and updates its evaluation round $R_{j+1}$ in a hashmap named $d_i$.last_try_round (Algo. 6: line 15), and $A_i^j$ will be chosen to perform local update (Algo. 6: line 17). Additionally, $d_i$ increments the retry interval of $d_{i,eval}^j$ in another hashmap $d_i$.retry_interval (Algo. 6: line 16), in which all of $d_i$'s candidate favorite neighbors are assigned an initial retry interval value of 0 (Algo. 1: line 13). The same pattern of retry interval incrementation occurs when $d_{i,eval}^j$ is tried again in subsequent rounds without being favored. The combination of

last_try_round and retry_interval forms the retry control mechanism for *NeighborFL*. This mechanism prevents a $cfn$ (i.e., a specific $d^j_{i,eval}$) from being frequently retried without considering other available candidate neighbors (Algo. 2, lines 5-8). This mechanism can be seen as a form of penalty that delays the future inclusion of a local model from an evaluated candidate that fails to contribute effectively to the timely prediction process. Note that the retry pattern (Algo. 6, lines 15-16) can be designed differently, such as decreasing the retry interval if $d^j_{i,eval}$ is re-added to $d_i.FN$ and remains there for an extended period.

The evaluation process of $A^j_{i,eval}$ by $d_i$ in round $R_{j+1}$, $j \geq 1$ is defined in Algo. 6. The output $A^j_{i,chosen}$ represents the chosen aggregated model to perform local update at the end of $R_j$ (Algo. 6, lines 13, 17). Algo. 6 also involves calculating the reputation for $d^j_{i,eval}$ (Algo. 6, lines 9-10), which will be used to illustrate a favorite neighbor removal method in Sec. 3.4.2.

In situations where $\mathcal{O} > 1$, handling the instances in $Y$, $P_{eval}$ and $P$ requires special consideration to align the error calculations correctly (Algo. 6, lines 7-8). Table. 1 illustrates the global sequence number of data points collected in $Y$ and the global sequence number of the predicted values in $P$ by a random device following Algorithm 5 for $\mathcal{I} = 12$, $\mathcal{O} = 1, 2, 3$ during $R_1$, $R_2$ and $R_3$, with $\tau^1 = 24$ and $\tau^{2+} = 12$. For example, when $\mathcal{O} = 2$ and during the 4th step (i.e., $m = 4$, Algo. 5, line 6) in $R_2$, according to Algo. 5, a device would first predict the 28th and 29th traffic values (Algo. 5, lines 10-12), denoted as the prediction instance $\hat{y}$ in green. Then, it collects the 28th true data point (Algo. 5, lines 13-14), colored in gray, and extracts the 27th and 28th true data points from $data^2$ (Algo. 5, line 16) denoted as the truth value instance $y$ in yellow. Note that $m = 1, 2, ..., 12$ in $R_1$ are omitted from the table as the devices haven't started training nor real-time prediction during those steps.

From the table, it can be inferred that for $\mathcal{O} = 2$, the sequences of the data points in the instances of $Y$ and $P$ differ by one position. Specifically, the first prediction instance $\hat{y}$ in $P$ corresponds to the 2nd truth instance $y$ in $Y$. Similarly, for $\mathcal{O} = 3$, the first $\hat{y}$ in $P$ corresponds to the 3rd $y$ in $Y$. In general, the corresponding $y$ in $Y$ of the $\hat{y}$ in $P$ is shifted by $\mathcal{O} - 1$ positions in terms of index.

To ensure that $d_i$ accommodates data collection shifts correctly while evaluating a candidate's local model $L^j_{i,try}$ in $R_{j+1}$, $j \geq 1$, one approach could be to let $A^j_{i,try}$ infer the latest $\mathcal{O} - 1$ predictions in $R_j$ using the historical data and append these $\mathcal{O} - 1$ $\hat{y}$'s to the beginning of $P^{j+1}_{i,eval}$, and also append the latest $\mathcal{O} - 1$ $\hat{y}$'s of $d_i$ to $P^{j+1}_i$. This way, the indexes of the sequences in $Y^{j+1}_i$, $P^{j+1}_{i,eval}$, and $P^{j+1}_i$ will align. A simpler alternative is to drop the first $\mathcal{O} - 1$ $y$'s in $Y^{j+1}_i$ and drop the last $\mathcal{O} - 1$ $\hat{y}$'s in both $P^{j+1}_{i,eval}$ and $P^{j+1}_i$ , as shown in pink in Table. 1 for examples with $\mathcal{O} = 2$ and $\mathcal{O} = 3$. Using this method, we can maintain the evaluation of $d^j_{i,eval}$ after $d_i$ finishes collecting new data during $R_{j+1}$. Note that by using this approach, $\tau^{2+}$ has to be at least $\mathcal{O}$.

## 3.4 Removal of a favorite neighbor

With evolving traffic conditions, including non-recurrent events, seasonal variations, or route detours, neighboring devices can undergo substantial changes in traffic dynamics. Consequently, the local models derived from these devices may no longer contribute effectively or may even adversely impact the real-time predictions made by a given device that opts to include them in its favorite neighbors set (*FN*). Thus, the removal of poor performing favorite neighbors from a device's *FN* becomes crucial in maintaining accurate real-time predictions. In this section, we introduce our design of the removal mechanism, which includes two parts: a removal trigger and a removal method. However, it should be noted that this design is presented as an example and alternative approaches can be explored when implementing *NeighborFL* in practice.

*3.4.1 Removal Trigger.* The removal trigger is responsible for determining when a device should initiate the removal action for one or more favorite neighbors ($fn$) from its favorite neighbors set (*FN*). In the current design,

**Algorithm 6:** EVALUATECANDIDATE() - $d_i$ determines to add a $d^j_{i,eval}$ to $d_i.FN$ in $R_{j+1}$, $j \geq 1$

1 **Input:** $d^j_{i,eval}$, $P^{j+1}_{i,eval}$, $P^{j+1}_i$, $Y^{j+1}_i$, $O$
2 **Output:** $A^{j+1}_{i,chosen}$
3 $P^{j+1}_{i,eval}$.DROPLATEST($O - 1$);
4 $P^{j+1}_i$.DROPLATEST($O - 1$);
5 **if** $O > 1$ **then**
6     $Y^{j+1}_i$.DROPEARLIEST($O - 1$);
7 $E^{j+1}_{i,eval} \leftarrow$ CALCERROR($P^{j+1}_{i,eval}$, $Y^{j+1}_i$);
8 $E^{j+1}_i \leftarrow$ CALCERROR($P^{j+1}_i$, $Y^{j+1}_i$);
9 $\Delta E \leftarrow E^{j+1}_i - E^{j+1}_{i,eval}$;
10 $d_i$.rep_book[$d^j_{i,eval}$] $\leftarrow d_i$.rep_book[$d^j_{i,eval}$] + $\Delta E$;
11 **if** $E^{j+1}_{i,eval} < E^{j+1}_i$ **then**
12     $d_i$.*FN*.ADD($d^j_{i,eval}$);
13     $A^{j+1}_{i,chosen} \leftarrow A^j_{i,eval}$;
14 **else**
15     $d_i$.last_try_round[$d^j_{i,eval}$] $\leftarrow R_{j+1}$;
16     $d_i$.retry_interval[$d^j_{i,eval}$] $\leftarrow d_i$.retry_interval[$d^j_{i,eval}$] + 1;
17     $A^{j+1}_{i,chosen} \leftarrow A^j_i$;

the trigger is activated when the prediction error $E$ of a device increases over a specified number of continuous rounds, denoted as $v$ (Algo. 9, line 3).

*3.4.2 Removal Method.* We propose two distinct removal methods in this work: "remove by reputation" and "remove the last added". In the "remove by reputation" method, each device maintains a hashmap called *rep_book* (Algo. 1, lines 6, 14; Algo. 6, line 10; Algo. 7) to track the subjective reputation of its candidate favorite neighbors ($cfn$'s). The reputation is determined by calculating the difference between $E$ and $E_{eval}$ when a $cfn$ is evaluated as $d^j_{i,eval}$ (Algo. 6, line 9). If the difference $\Delta E = E - E_{eval}$ is a positive value, indicating that the local model of $d^j_{i,eval}$ ($L^j_{i,eval}$) contributes to reducing the real-time prediction error, $d^j_{i,eval}$ is assigned a positive reputation. Conversely, if $\Delta E$ is negative, $d^j_{i,eval}$ receives a negative reputation (Algo. 6, line 10). The reputation accumulates over the course of communication rounds when a $cfn$ is being (re-)evaluated. When a device removes a $fn$, it eliminates the $fn$ associated with the lowest reputation (Algo. 7, lines 6-8). Algo. 7 illustrates the process of "remove by reputation" executed by $d_i$. A removed $fn$ is denoted as ~$fn$.

The "remove the last added" method allows a device to remove $fn$ in the reverse order of their addition, effectively undoing the addition of a $fn$. As a device typically (but not definitely) adds $cfn$ to *FN* from closest to farthest, the device usually removes $fn$ from farthest to closest (Algo. 8: lines 4-5), following a similar heuristic as Algo. 2. In this method, a device keeps track of the order in which $cfn$'s are added to its *FN*, and always removes the most recently added $fn$. In this method, the *FN* data structure utilizes a stack instead of a set. This method intuitively exhibits a more chronological adaptation to time-dependent changes compared to the "remove by reputation" method. Algo. 8 outlines the process of the "remove the last added" method performed by $d_i$, assuming the removal of one $fn$ at a time.

Table 1. Global data point sequence of the instances in $Y$ and $P$ for $\mathcal{I} = 12, \mathcal{O} = 1, 2, 3$ during $R_1$, $R_2$ and $R_3$

| $\mathcal{R}$ | $m$ | Collect | $\mathcal{O} = 1$ | | $\mathcal{O} = 2$ | | $\mathcal{O} = 3$ | |
|---|---|---|---|---|---|---|---|---|
| | | | $y$ | $\hat{y}$ | $y$ | $\hat{y}$ | $y$ | $\hat{y}$ |
| Round 1 | 13 | 13th | 13 | 13 | NULL | 13, 14 | NULL | 13, 14, 15 |
| | 14 | 14th | 14 | 14 | 13, 14 | 14, 15 | NULL | 14, 15, 16 |
| | 15 | 15th | 15 | 15 | 14, 15 | 15, 16 | 13, 14, 15 | 15, 16, 17 |
| | 16 | 16th | 16 | 16 | 15, 16 | 16, 17 | 14, 15, 16 | 16, 17, 18 |
| | 17 | 17th | 17 | 17 | 16, 17 | 17, 18 | 15, 16, 17 | 17, 18, 19 |
| | 18 | 18th | 18 | 18 | 17, 18 | 18, 19 | 16, 17, 18 | 18, 19, 20 |
| | 19 | 19th | 19 | 19 | 18, 19 | 19, 20 | 17, 18, 19 | 19, 20, 21 |
| | 20 | 20th | 20 | 20 | 19, 20 | 20, 21 | 18, 19, 20 | 20, 21, 22 |
| | 21 | 21st | 21 | 21 | 20, 21 | 21, 22 | 19, 20, 21 | 21, 22, 23 |
| | 22 | 22nd | 22 | 22 | 21, 22 | 22, 23 | 20, 21, 22 | 22, 23, 24 |
| | 23 | 23rd | 23 | 23 | 22, 23 | 23, 24 | 21, 22, 23 | 23, 24, 25 |
| | 24 | 24th | 24 | 24 | 23, 24 | 24, 25 | 22, 23, 24 | 24, 25, 26 |
| Round 2 | 1 | 25th | 25 | 25 | 24, 25 | 25, 26 | 23, 24, 25 | 25, 26, 27 |
| | 2 | 26th | 26 | 26 | 25, 26 | 26, 27 | 24, 25, 26 | 26, 27, 28 |
| | 3 | 27th | 27 | 27 | 26, 27 | 27, 28 | 25, 26, 27 | 27, 28, 29 |
| | 4 | 28th | 28 | 28 | 27, 28 | 28, 29 | 26, 27, 28 | 28, 29, 30 |
| | 5 | 29th | 29 | 29 | 28, 29 | 29, 30 | 27, 28, 29 | 29, 30, 31 |
| | 6 | 30th | 30 | 30 | 29, 30 | 30, 31 | 28, 29, 30 | 30, 31, 32 |
| | 7 | 31th | 31 | 31 | 30, 31 | 31, 32 | 29, 30, 31 | 31, 32, 33 |
| | 8 | 32th | 32 | 32 | 31, 32 | 32, 33 | 30, 31, 32 | 32, 33, 34 |
| | 9 | 33th | 33 | 33 | 32, 33 | 33, 34 | 31, 32, 33 | 33, 34, 35 |
| | 10 | 34th | 34 | 34 | 33, 34 | 34, 35 | 32, 33, 34 | 34, 35, 36 |
| | 11 | 35th | 35 | 35 | 34, 35 | 35, 36 | 33, 34, 35 | 35, 36, 37 |
| | 12 | 36th | 36 | 36 | 35, 36 | 36, 37 | 34, 35, 36 | 36, 37, 38 |
| Round 3 | 1 | 37th | 37 | 37 | 36, 37 | 37, 38 | 35, 36, 37 | 37, 38, 39 |
| | 2 | 38th | 38 | 38 | 37, 38 | 38, 39 | 36, 37, 38 | 38, 39, 40 |
| | 3 | 39th | 39 | 39 | 38, 39 | 39, 40 | 37, 38, 39 | 39, 40, 41 |
| | 4 | 40th | 40 | 40 | 39, 40 | 40, 41 | 38, 39, 40 | 40, 41, 42 |
| | 5 | 41st | 41 | 41 | 40, 41 | 41, 42 | 39, 40, 41 | 41, 42, 43 |
| | 6 | 42nd | 42 | 42 | 41, 42 | 42, 43 | 40, 41, 42 | 42, 43, 44 |
| | 7 | 43rd | 43 | 43 | 42, 43 | 43, 44 | 41, 42, 43 | 43, 44, 45 |
| | 8 | 44th | 44 | 44 | 43, 44 | 44, 45 | 42, 43, 44 | 44, 45, 46 |
| | 9 | 45th | 45 | 45 | 44, 45 | 45, 46 | 43, 44, 45 | 45, 46, 47 |
| | 10 | 46th | 46 | 46 | 45, 46 | 46, 47 | 44, 45, 46 | 46, 47, 48 |
| | 11 | 47th | 47 | 47 | 46, 47 | 47, 48 | 45, 46, 47 | 47, 48, 49 |
| | 12 | 48th | 48 | 48 | 47, 48 | 48, 49 | 46, 47, 48 | 48, 49, 50 |

Lastly, to avoid rapid re-evaluation and addition of a $\sim fn$ in subsequent communication rounds, a device must maintain a record of $\sim fn$ in its last_try_round and retry_interval hashmap (Algo. 9, lines 5-6). The complete favorite neighbor removal steps are described in Algo. 9.

**Algorithm 7:** REMOVEBYREP() - $d_i$ removes a $fn$ by reputation

**Input:** None;<br>
**Output:** $\sim fn$<br>
SORT($d_i$.*rep_book*);<br>
/* sort $d_i$'s rep_book by reputation value from low to high */<br>
**for** *each* $cfn$ *in* $d_i$.*rep_book* **do**<br>
    **if** $cfn \in d_i$.FN **then**<br>
        $d_i$.*FN*.REMOVE($cfn$);<br>
        $\sim fn \leftarrow cfn$;<br>
        exit;

**Algorithm 8:** REMOVELASTADDED() - $d_i$ removes a $fn$ by reverse order of addition

**Input:** None;<br>
**Output:** $\sim fn$;<br>
**if** $d_i$.FN *not empty* **then**<br>
    $\sim fn \leftarrow d_i$.*FN*.POP();<br>
    exit;

**Algorithm 9:** REMOVEFAVORITENEIGHBOR() - $d_i$ removes a $fn$ from its *FN* set in $R_j$, $j > v$

**Input:** Trigger rounds value $v$;<br>
**Output:** None;<br>
**if** $E_i^j > E_i^{j-1} > ... > E_i^{j-v+1}$ **then**<br>
    /* Removal trigger - error $E_i$ has been continuously increasing for $v$ rounds */<br>
    $\sim fn \leftarrow d_i$.REMOVENEIGHBOR();<br>
    /* Algo. 7 or 8 */<br>
    $d_i$.last_try_round[$\sim fn$] $\leftarrow R_j$;<br>
    $d_i$.retry_interval[$\sim fn$] $\leftarrow d_i$.retry_interval[$\sim fn$] + 1;

### 3.5 NeighborFL complete operations

The preceding subsections have divided *NeighborFL* into various modules. This section consolidates those operations and presents the overall top-to-bottom functioning of *NeighborFL* in Algo. 10. Along with forming the *CFN* hash map, a device also initializes five variables (Algo. 10, lines 5-9). The initial model provided to the devices in $\mathcal{D}$ is denoted by $A^0$, which can be randomly generated or pretrained. Different devices may have different initial models, and thus $A_i^0$ denotes the initial model of $d_i$ before it participates in *NeighborFL* (Algo. 10, line 8).

The first three actions a device performs upon starting a new communication round are to make two predictions while collecting the incoming data points to update its local dataset (Algo. 10: lines 13-16). It is important to underscore that these three functions operate concurrently in an asynchronous manner, governed by the shared time stamp variable $m$ (Algo. 3, lines 3, 5, 8; Algo. 5, lines 3, 6, 18), as well as $\tau$. Accompanied Fig. 3 visualizes these operations with $\mathcal{I} = 12$, $\mathcal{O} = 1$, and $\tau^j = 12$, showing for $m = 1, \ldots, 12$ how $d_i$ concurrently updates $data_i^j$ with each arriving $x_{i,m}^{j}$, produces $\hat{y}$ and $\hat{y}_{eval}$ via $A_i^{j-1}$ and $A_{i,eval}^{j-1}$, and depicts the evolution of $P_i^j$, $P_{i,eval}^j$, and $Y_i^j$ throughout communication round $R_j$, $j > 1$.

After completing all $\tau$ steps, at the end of a communication round, the device executes the end-of-round pipeline: (i) if a candidate favorite neighbor was chosen, it evaluates the candidate's local model; (ii) it performs local training; (iii) it aggregates local models; and (iv) it may remove one or more favorite neighbors and select one or more new candidates for evaluation in the next round. Fig. 4 illustrates the candidate evaluation and local training for $d_i$ in $R_j$, $j > 1$. The AGGREGATE() function takes a device's freshly produced local model and the local models from its *Favorite Neighbors* to perform model aggregation and generate $A$ (Algo. 10, line 21), typically using a method like *Federated Averaging* [19]. If an available candidate is selected, the same aggregation method should also be used to produce $A_{eval}$ (Algo. 10, lines 24-25). Also note that the removal action should occur before a device selects an available candidate (Algo. 10, lines 22-23). This is done to prevent a scenario where a newly selected candidate has the lowest reputation and gets removed immediately after being chosen, particularly when a device adopts the "remove by reputation" method.

## 4 Experimental Results

### 4.1 Experimental Design

*4.1.1 Dataset.* We manually selected a study region from the widely available reference dataset, PEMS-BAY[12], collected by the California Transportation Agency Performance Measurement System (PeMS). Our chosen area covers 26 devices from the total 325 devices within the dataset. In Fig. 5, the locations of all 325 devices are shown in the lower left corner. The devices included in our experiments are enclosed within the red region, and their corresponding device IDs are labeled next to the devices' locations in the remainder of the figure, showing a zoomed-in map. Each device's ID includes an underscore followed by the letter "N" or "S" to indicate the traffic direction it monitors, either North or South. Without loss of generality, we use traffic speed as the modeling variable in this study.

*4.1.2 Deep Learning Model.* In our experiments, we employed a light-weight construction of the Long Short-Term Memory (LSTM) [10] variety of recurrent neural network (RNN) because it can be efficiently trained on heterogeneous edge devices for FLTP applications [8, 14, 20–22, 24]. Following the convention of federated learning, all devices utilized identical LSTM networks. These networks consisted of 12 input neurons ($\mathcal{I} = 12$), corresponding to one hour of traffic data with a 5-minute time resolution, and one output neuron ($\mathcal{O} = 1$), enabling a one-step 5-minute look-ahead prediction, as discussed in Sec. 3.3.2. The LSTM network is constructed with two hidden layers of size 128 hidden units. Furthermore, the two LSTM layers are followed by a dropout layer initialized at 0.2 and a fully connected output layer of size $\mathcal{O} = 1$.

*4.1.3 Baseline Methods.* To assess the performance of *NeighborFL*, we conducted a comparative analysis with three baseline methods: *Centralized*, *NaiveFL*, and *r-NaiveFL*. These baseline methods are also streaming methods and exhibit the following distinctions compared to *NeighborFL*:

a) The *Centralized* model of a device undergoes continuous real-time training without federation at the end of each communication round. From this perspective, the *Central* model trained by a device can be seen as updated solely by this specific device's incoming data throughout the entire process. This model was initially introduced in [20], and its performance highlighted here serves to emphasize the benefits of adopting federated learning for traffic prediction. Technically, we can interpret the training process of a *Central* model as a device performing *NeighborFL* where the *Faviorite Neighbors* set remains empty indefinitely.
b) The *NaiveFL* model is also updated online in real-time and follows the conventional FL aggregation method. In every communication round, all devices share the same aggregated (global) model for prediction and

**Algorithm 10:** NeighborFL() - Overall Operations of *NeighborFL*.

**Input:** $\boldsymbol{A}^0$, $\mathcal{D}$, $\boldsymbol{r}$, $\boldsymbol{\tau}$, $\mathcal{I}$, $\mathcal{O}$, $MaxDataSize$, $\mathcal{E}$, $v$;
**Output:** None;
**for** *each device* $d_i \in \mathcal{D}$ *in parallel* **do**
  $d_i$.FormCFN($r_i$) ; // Algo. 1
  $d_i.FN \leftarrow []$;
  $data_i^0 \leftarrow []$;
  $d_{i,eval}^0 \leftarrow$ None;
  $A_i^0 \leftarrow A_i^0 \in \boldsymbol{A}^0$;
  $A_{i,eval}^0 \leftarrow$ None;
**for** *communication round* $R_j$ *in* $\{R_1, R_2, ...\}$ **do**
  **for** *each device* $d_i \in \mathcal{D}$ *in parallel* **do**
    /* Below 3 functions run concurrently. */
    $data_i^j \leftarrow d_i$.UpdateDataset($data_i^{j-1}$, $\tau^j$, $MaxDataSize$) ; // Algo. 3
    $P_i^j, Y_i^j \leftarrow d_i$.Predict($A_i^{j-1}$, $data_i^j$, $\tau^j$, $\mathcal{I}$, $\mathcal{O}$) ; // Algo. 5
    **if** $d_{i,eval}^{j-1}$ *is not None* **then**
      $P_{i,eval}^j, Y_i^j \leftarrow d_i$.Predict($A_{i,eval}^{j-1}$, $data_i^{j-1}$, $\tau^j$, $\mathcal{I}$, $\mathcal{O}$) ; // Algo. 5
    /* Above 3 functions run concurrently. */
    $A_{i,chosen}^j \leftarrow A_i^{j-1}$ ; // By default
    **if** $d_{i,eval}^{j-1}$ *is not None* **then**
      $A_{i,chosen}^j \leftarrow d_i$.EvaluateCandidate($d_{i,eval}^{j-1}$, $P_{i,eval}^j$, $P_i^j$, $Y_i^j$, $\mathcal{O}$) ; // Algo. 6
    $L_i^j \leftarrow d_i$.LocalTraining($A_{i,chosen}^j$, $data_i^j$, $\mathcal{E}$, $\mathcal{I}$, $\mathcal{O}$) ; // Algo. 4
    $A_i^j \leftarrow$ Aggregate($L_i^j$, $d_i.FN$) ; // e.g., FedAvg
    $d_i$.RemoveFavoriteNeighbor($v$) ; // Algo. 9
    $d_{i,eval}^j \leftarrow d_i$.SelectCandidate() ; // Algo. 2
    **if** $d_{i,eval}^j$ *is not None* **then**
      $A_{i,eval}^j \leftarrow$ Aggregate($L_i^j$, $L_{i,eval}^j$, $d_i.FN$) ; // e.g., FedAvg

learning. It serves as the primary model for comparison with *NeighborFL*, emphasizing the effectiveness of the aggregation heuristics employed in *NeighborFL*. Note that the conventional FL approach, represented on the left side of Fig. 1, illustrates the workflow of *NaiveFL*. Formally, we can think of the training process of a *NaiveFL* model in this context as a device implementing *NeighborFL* where its *Favorite Neighbors* set includes the entire cohort of devices in the study region, excluding itself, from the onset of the FL process.

c) The *r-NaiveFL* model can be viewed as a hybrid between *NeighborFL* and *NaiveFL*. Essentially, the training process of a *r-NaiveFL* model, when performed by a device, involves filling its *Favorite Neighbors* set with devices within its radius $r$ (used to form its *Candidate Favorite Neighbors* set), excluding itself, at initialization. Notably, unlike *NaiveFL*, *r-NaiveFL* still enables each device to have a customized aggregated model for training and prediction in each communication round. As both *r-NaiveFL* and *NeighborFL* select devices from the same area for federation, with the only difference being that *NeighborFL* dynamically adjusts a device's *Favorite Neighbors* set, we present the performance of *r-NaiveFL* to further highlight the effectiveness of the heuristics employed in *NeighborFL* and the benefits of evolutionary *Favorite Neighbors* sets.

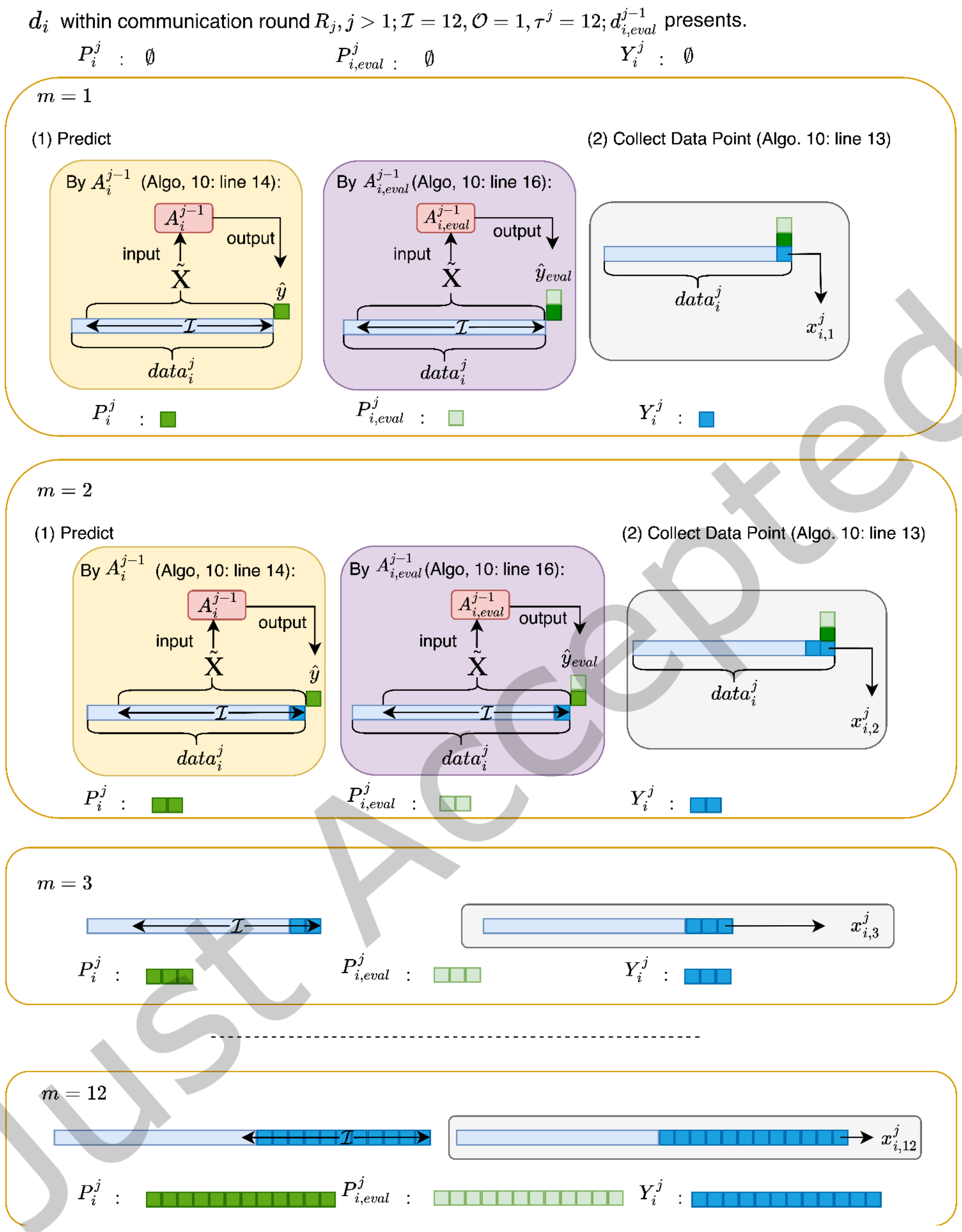

Fig. 3. Visualization of $d_i$ performing dataset updates and predictions in round $R_j, j > 1$, with $\mathcal{I} = 12$, $\mathcal{O} = 1$, $\tau^j = 12$, and the presence of $d_{i,eval}^{j-1}$.

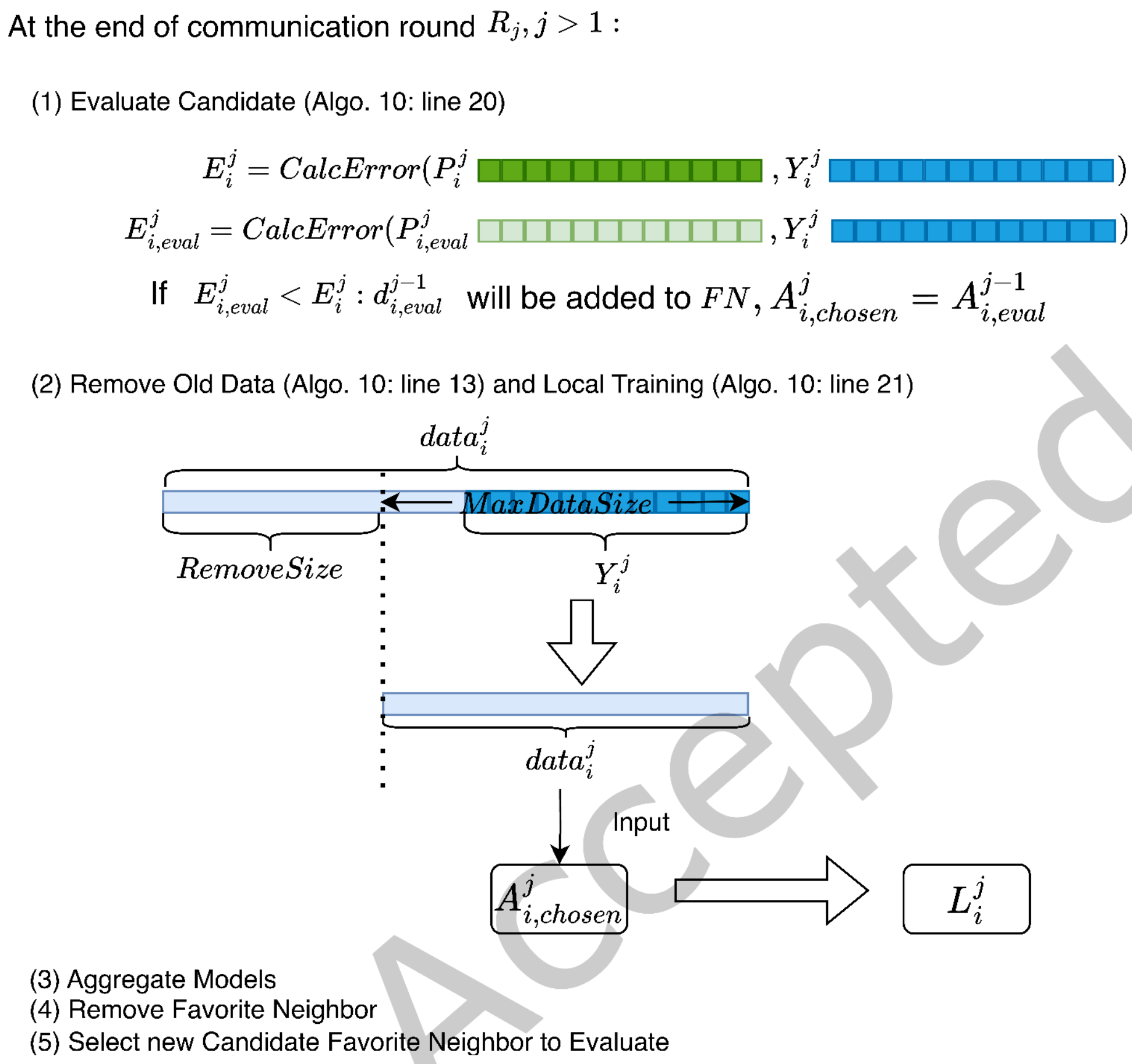

Fig. 4. End-of-round workflow for $d_i$ in round $R_j, j > 1$: candidate evaluation and local training. Visualizations of the remaining operations are omitted.

For all four learning methods, including *NeighborFL*, we utilized an identical LSTM network architecture outlined in Sec. 4.1.2. The optimizer chosen for all methods was RMSProp [9]. In each communication round, all methods used a local epoch of 5 ($\mathcal{E}$ = 5) and a batch size of 1.

*4.1.4 Number of Predictions $\tau$.* In most traffic datasets, including PEMS-BAY [4] used in this study, the time resolution is set to 5 minutes. To align the communication round with hours, we set $\tau^{2+}$ to 12 for $R_2$ and subsequent rounds. This indicates that a device needs to collect 12 data points and makes 12 one-step look-ahead predictions in $R_{2+}$, corresponding to one hour of data in the PEMS-BAY dataset. However, because the LSTM models employed in this study have 12 input neurons and a single output neuron, they require a minimum of $I + O = 13$ data points to form the online training batch and begin learning. Consequently, we set $\tau^1$ to 24 to ensure $\tau^1 \geq I + O$ is satisfied in $R_1$, which is equivalent to 2 hours of data. This configuration allows devices to make 24 - 12 - 1 + 1 = 12 predictions in $R_1$ (Algo. 5: lines 5-8, 10-12), as calculated by Eq. 2, thereby unifying the number of predictions for all communication rounds to be 12.

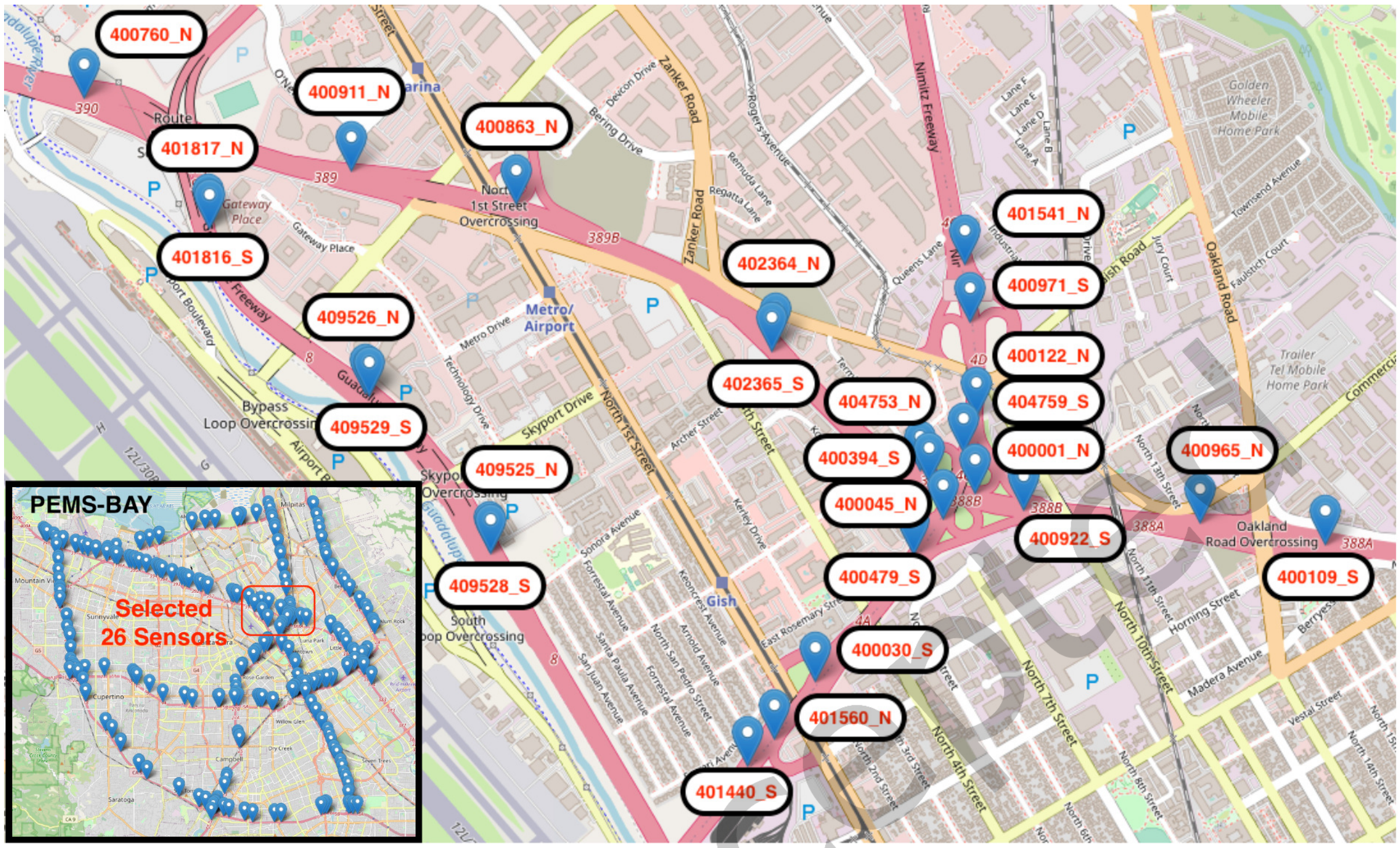

Fig. 5. Map of PEMS-BAY and the study region containing our 26 selected devices for experiments.

*4.1.5 NeighborFL Setup.* For the *NeighborFL* method, we conducted experiments using four different favorite neighbor removal strategies to assess their impact on prediction performance. These strategies were compared while keeping other parameters consistent. The four strategies are summarized as follows:

(1) Remove-by-reputation (Algo. 7) with a removal trigger of $v = 1$ (referred to as R1).
(2) Remove-by-reputation (Algo. 7) with a removal trigger of $v = 3$ (referred to as R3).
(3) Remove-the-last-added (Algo. 8) with a removal trigger of $v = 1$ (referred to as L1).
(4) Remove-the-last-added (Algo. 8) with a removal trigger of $v = 3$ (referred to as L3).

Furthermore, we used a fixed radius ($r = 1$ mile) for all devices to form their *Candidate Favorite Neighbors Set* (*CFN*). The number of candidate favorite neighbors (#*CFN*) for each device is listed in Table 2.

Table 2. Number of *CFN* for each device

| Device ID | #*CFN* | Device ID | #*CFN* |
|---|---|---|---|
| 401816_S | 8 | 401817_N | 8 |
| 400911_N | 10 | 400863_N | 18 |
| 409526_N | 13 | 409529_S | 13 |
| 409525_N | 21 | 409528_S | 21 |
| 402364_N | 21 | 402365_S | 21 |
| 401541_N | 16 | 400971_S | 19 |
| 400122_N | 19 | 404759_S | 19 |
| 400394_S | 19 | 404753_N | 19 |
| 400045_N | 19 | 400001_N | 18 |
| 400922_S | 18 | 400479_S | 19 |
| 400030_S | 20 | 401560_N | 19 |
| 401440_S | 18 | 400965_N | 16 |
| 400109_S | 12 | 400760_N | 6 |

For example, device *400760_N* has 6 candidate favorite neighbors (*401817_N*, *400911_N*, *401816_S*, *400863_N*, *409526_N*, *409529_S*) within a 1-mile radius, which is consistent with Fig. 5.

Additionally, we set *MaxDataSize = 72* for all devices, which corresponds to 6 hours of traffic data. The Mean Squared Error (MSE) was used to evaluate the real-time prediction error of a candidate neighbor's local model (Algo. 6, lines 9-10), as we want a device to skip a candidate neighbor's model producing large errors. Moreover, as pretrained models have previously been utilized to initialize the models used in FLTP applications [33, 35], we have also evaluated the effects of pretraining on the performance of *NeighborFL* and the three baseline methods. In the experiments involving pretrained models, we used the first week of traffic speed data from PEMS-BAY (ranging from 2017-01-01 00:00:00 to 2017-01-07 23:55:00, both timestamps included) as training data for each of the 26 sensors. We trained 26 models individually using an LSTM network with randomly initialized weights (represented as $A^0$) with the structure mentioned in 4.1.2, applying 5 local epochs and a batch number of 1. The resulting trained models are denoted as $A_1^0$, $A_2^0$, ..., $A_{26}^0$ and serve as the initial models for each corresponding device in real-time federated learning. For experiments involving non-pretrained models, each device uses $A^0$ as its initial model.

Our simulations were performed on Google Colab using an NVIDIA Tesla T4 GPU with standard RAM. The files containing the experimental results mentioned in Sec. 4.1.6 are available in our GitHub repository and can be reproduced using a random seed value of 40.

*4.1.6 Results.* We conducted a total of 250 communication rounds across 26 devices for all of our experiments, which involved both pretrained and non-pretrained initial models. Within each of these two running types, we employed three baseline methods and four different configurations of *NeighborFL* (R1, R3, L1, L3), resulting in a total of 14 running methods, as indicated in Table. 3 and Table. 5. These 250 rounds utilized speed data for training from 2017-01-08 00:00:00 to 2017-01-18 10:55:00 (both timestamps inclusive), and the devices provided real-time predictions for speed values from 2017-01-08 01:00:00 to 2017-01-18 10:55:00 (both timestamps inclusive).

To visually compare the real-time prediction accuracy among these methods, we have plotted the real-time prediction curves (i.e., values in $P_{fav}$, Algo. 10, line 13) alongside the ground truth curves (i.e., values in $Y$, Algo. 10, line 13) for the last 24 hours of the simulated federation (i.e., rounds 227 to 250). Due to space constraints, we presented the curves from three baseline and the *NeighborFL L1* methods under the pretrained setting, as shown in Fig. 6. The criteria for choosing these four out of the 14 methods are referred to Table. 3, which outlines the **average device** mean squared error (MSE) across all 26 devices for the last 24 rounds between the ground truth values and the corresponding methods.

As an example of the calculation of the average device MSE values for any of these 14 methods shown in Table. 3, the Pretrain MSE value of *NeighborFL L1* (i.e., 7.45 highlighted in pink in Table. 3) is computed as follows:

(1) First, we calculate the MSE value between the *NeighborFL L1* predictions and the corresponding ground truth values from rounds 227 to 250 for each of the 26 devices. For a random device $d_i$, the MSE value is obtained by following Eq. 3,

$$MSE_i^{m\sim n} = \frac{1}{N} \sum_{k=1}^{N} \left( \hat{y}^{(k)} - y^{(k)} \right)^2 \tag{3}$$

where

$$P_i^{m\sim n} = P_i^m \cup P_i^{m+1} \cup \ldots \cup P_i^n, m \leq n$$
$$Y_i^{m\sim n} = Y_i^m \cup Y_i^{m+1} \cup \ldots \cup Y_i^n, m \leq n$$
$$N = |P_i^{m\sim n}| = |Y_i^{m\sim n}| = (m - n + 1) * \tau^{2+}$$
$$\hat{y}^{(k)} \in P_i^{m\sim n} : \text{the } k\text{-th prediction instance in } P_i^{m\sim n}$$
$$y^{(k)} \in Y_i^{m\sim n} : \text{the } k\text{-th ground truth instance in } Y_i^{m\sim n}$$
$$MSE_i^{m\sim n} : \text{MSE value of } d_i \text{ from round } m \text{ to } n$$

In this case, $m = 227$, $n = 250$, $\tau^{2+} = 12$; $P_i^{m\sim n}$ corresponds to $P_i^{227\sim 250}$ of *NeighborFL L1*, and $Y_i^{m\sim n}$ represents the truth values $Y_i^{227\sim 250}$ extracted from the raw PEMS-BAY dataset [12] of $d_i$ with the associated device ID for the corresponding time range. The resulting MSE values for each of the 26 devices are listed under the *NeighborFL L1* column in Table. 4.

(2) These individual device MSE values are then summed up and divided by the number of devices in $\mathcal{D}$ to obtain the average device MSE value, as shown in Eq. 4.

$$AVGMSE^{m\sim n} = \frac{1}{|\mathcal{D}|} \sum_{i=1}^{|\mathcal{D}|} MSE_i^{m\sim n} \tag{4}$$

In this case, $|\mathcal{D}| = 26$, and the resulting value $AVGMSE^{227\sim 250}$ for *NeighborFL L1* is 7.45, serving as an overall last 24 hours error evaluation metric. The same calculation procedure is applied to generate the other 13 average MSE values presented in Table 3.

From Table. 3, we observe that *NeighborFL L1* had the smallest Average Device MSE value (highlighted in pink) among all 14 methods. In the pretrained setting, all four *NeighborFL* configurations exhibited lower average MSE values compared to the three baseline methods, highlighting the superior real-time learning and prediction capabilities of *NeighborFL*. For instance, *NeighborFL L1* reduces the MSE value by 16.9% compared to *NaiveFL*. A similar trend is observed in the non-pretrained setting, where *NeighborFL L3* achieves the lowest average MSE value (highlighted in green) among all methods. Although the MSE value of *r-NaiveFL* (i.e., 10.88) is lower than that of *NeighborFL L1* and *NeighborFL R3*, the difference is marginal, and all four *NeighborFL* configurations outperformed *NaiveFL* by at least 5%, while also significantly outperforming *Central* under the non-pretrained setting. Furthermore, by comparing the Average Device MSE values of the seven methods between the pretrained and non-pretrained settings, we observed that using pretrained models can significantly enhance the training and prediction performance for all seven configurations. Specifically, for *NeighborFL*, pretraining the initial models with one week of historical data can reduce the prediction error by at least 30%, as indicated in the table.

Table 3. Average Device MSE of last 24 rounds

| Methods | Pretrain MSE | Non-Pretrain MSE |
|---|---|---|
| Central | 13.14 | 19.53 |
| NaiveFL | 8.97 | 11.01 |
| r-NaiveFL | 8.35 | 10.88 |
| NeighborFL L1 | 7.45 | 10.9 |
| NeighborFL L3 | 7.83 | 10.38 |
| NeighborFL R1 | 7.8 | 10.5 |
| NeighborFL R3 | 7.74 | 10.89 |

In Fig. 6, the curves representing the speed predictions produced by *Central*, *NaiveFL*, and *r-NaiveFL* methods are shown in orange, green, and gray, respectively. The curve of the *NeighborFL L1* method is depicted in red, while the ground truth curve is displayed in blue. The x-axis represents the communication round, with each round consisting of 12 steps of speed measurements between any two round indexes, since we chose $O = 1$ and $\tau^{2+} = 12$. By closely examining the curves of the representative device *400760_N*, we can observe that the *NeighborFL* method consistently achieves better overall predictions compared to the three baseline methods. The red curve is generally closer to the blue curve compared to the other three curves, which is particularly evident during rounds 246 to 250. Moreover, *NeighborFL* demonstrates enhanced adaptability to sudden speed changes detected by the device, as evidenced by the 12 speed steps in the round 246 to 247 segment. This trend is also observed in other devices such as *401817_N*, *400911_N*, *409529_S*, and *400922_S*.

To further assess the performance of *NeighborFL* at an individual device level, we provide the MSE values for each of the 26 devices over the last 24 rounds, categorized by the different methods, in Table 4. In each device's row, the lowest MSE value is highlighted in green. By counting the number of green highlights under the *NeighborFL L1* column, we observe that out of the 26 devices, *NeighborFL* outperforms all three baseline methods in 18 devices. For the remaining 8 devices, 5 devices (i.e., *400394_S, 400045_N, 400001_N, 401560_N, 400109_S*) still have their *NeighborFL* outperform *NaiveFL* method. Overall, *NeighborFL* surpasses *NaiveFL* in 88% of the devices within the entire cohort during the last 24 hours of the simulations.

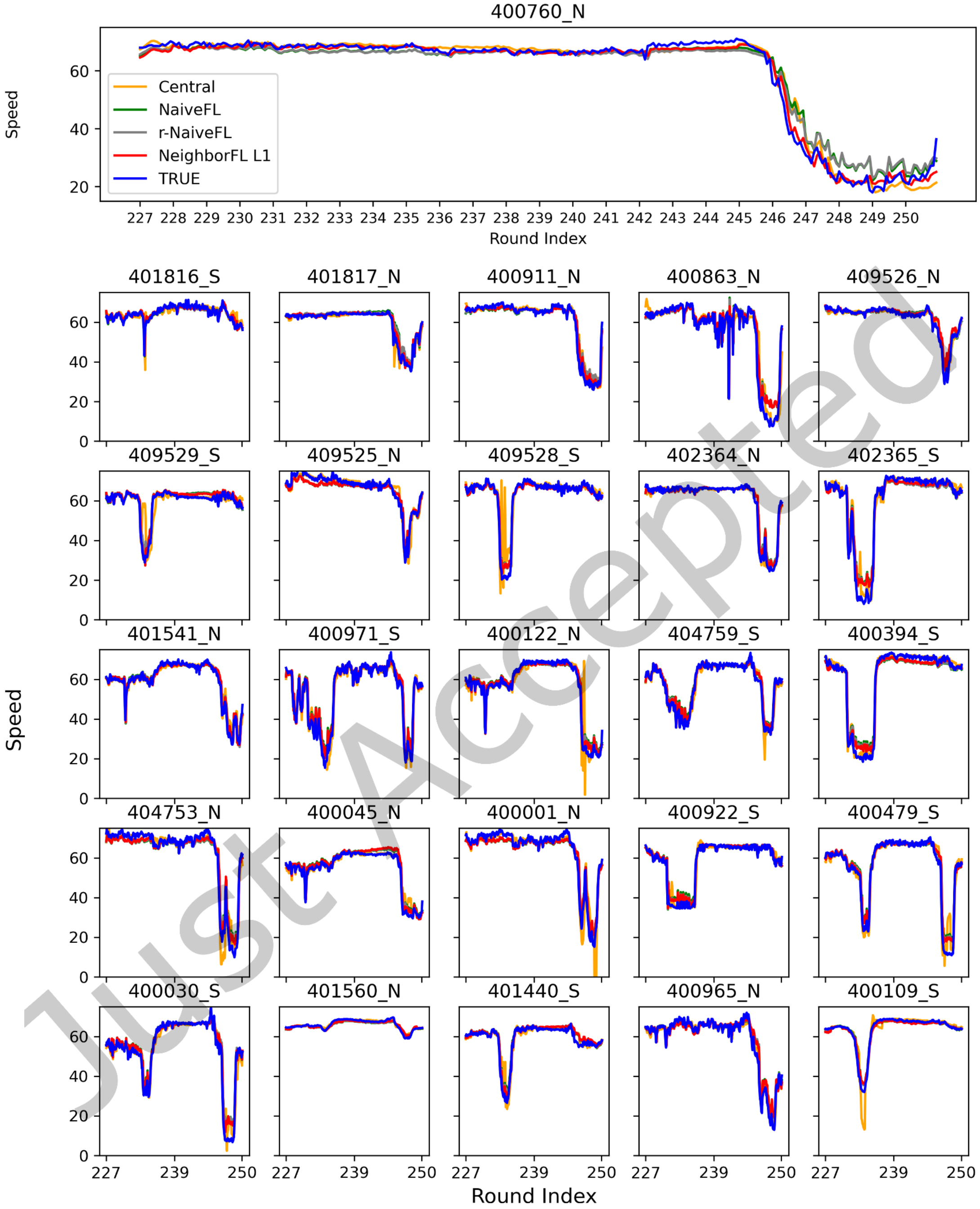

Fig. 6. Prediction curves of 26 devices in the last 24 rounds among baseline methods and *NeighborFL L1*

Table 4. Individual Device MSE Values in last 24 rounds

| | Central | NaiveFL | r-NaiveFL | NeighborFL L1 |
|---|---|---|---|---|
| 401816_S | 6.81 | 2.69 | 2.98 | 3.14 |
| 401817_N | 4.59 | 5.36 | 4.88 | 2.65 |
| 400911_N | 10.56 | 6.81 | 7.47 | 5.31 |
| 400863_N | 41.67 | 33.61 | 32.9 | 31.31 |
| 409526_N | 4.29 | 3.76 | 3.47 | 6.55 |
| 409529_S | 24.14 | 4.51 | 5.01 | 3.57 |
| 409525_N | 8.03 | 7.43 | 7.41 | 7.08 |
| 409528_S | 42.1 | 5.4 | 5.1 | 5.09 |
| 402364_N | 5.18 | 6.03 | 5.79 | 5 |
| 402365_S | 15.43 | 17.24 | 15.75 | 13.46 |
| 401541_N | 6.4 | 4.58 | 4.35 | 4.3 |
| 400971_S | 14.74 | 14.77 | 13.21 | 13.01 |
| 400122_N | 22.8 | 7.78 | 6.69 | 5.73 |
| 404759_S | 8.27 | 7.19 | 6.37 | 5.99 |
| 400394_S | 5.52 | 13.14 | 10.68 | 8.6 |
| 404753_N | 18.9 | 14.65 | 13.24 | 11.16 |
| 400045_N | 5.99 | 5.66 | 5.35 | 5.47 |
| 400001_N | 21.36 | 10.14 | 9.41 | 9.6 |
| 400922_S | 7.08 | 6.56 | 5.42 | 4.96 |
| 400479_S | 16.59 | 12.09 | 10.21 | 9.84 |
| 400030_S | 13.93 | 15.67 | 15.09 | 11.69 |
| 401560_N | 0.47 | 0.76 | 0.74 | 0.66 |
| 401440_S | 9.37 | 4.39 | 4.03 | 3.67 |
| 400965_N | 8.08 | 10.25 | 9.98 | 10.63 |
| 400109_S | 12.19 | 2.02 | 1.27 | 1.94 |
| 400760_N | 7.09 | 10.75 | 10.23 | 3.22 |

Another way to visualize the performance of *NeighborFL* compared to the three baselines is to examine the overall trend of MSE throughout the entire simulation (i.e., from communication round 1 to 250). In Fig. 7, we present how the MSE values of the predictions generated by the three baselines and *NeighborFL L1* change in relation to the ground truth for each of the 26 devices as the communication round progresses. The color scheme for the four methods follows that of Fig. 6. Unlike Fig. 6, the y-axis represents the MSE value, and the x-axis represents round ranges instead of individual round indexes. Each round range represents a group of round indexes, such as Round Range 24-48 indicating communication rounds 24, 25, 26, ..., 48 (inclusive). We group the error values into round ranges to fit the curves within the small subfigures while preserving the true trends. Each round range consists of 24 communication rounds, except for the first round range which has 23 rounds (due to the absence of predictions in the initial hour of the simulation), and the last round range which has 10 rounds. The calculation of these MSE values follows the same formula as Eq. 3.

Each plot also includes three percentage values displayed at the top. These percentages indicate the number of times *NeighborFL L1* achieves a lower MSE value than the corresponding baseline method across the 11 smoothed MSE values within the respective round ranges on the x-axis, for a particular device. For example, for device *409529_S*, *NeighborFL L1* produces MSE values that are lower than *Central* 6 times and lower than both *NaiveFL* and *r-NaiveFL* 8 times, resulting in percentages of 6/11=54.55% and 8/11=72.73%, respectively. When a percentage value exceeds 50%, the corresponding method and the percentage are highlighted in red, indicating that *NeighborFL L1* outperforms the particular baseline method for the majority of the simulation. Notably, for devices where *NeighborFL L1* outperforms *NaiveFL* by more than 50%, a star is displayed in the upper left corner of the figure.

The device *400922_S* has been selected as the representative for this analysis. By examining the error curves of *400922_S*, we can clearly observe that the error of *NeighborFL L1* consistently remains lower than the three baseline methods, as indicated by the highlighted red values. Among the remaining 25 devices, we have seen that 22 devices outperform *NaiveFL* according to the stars, further highlighting the effectiveness of *NeighborFL* compared to *NaiveFL*.

While it is noticeable that the error of the *Central* method exhibits greater fluctuations in certain devices such as *401816_S*, *400911_N*, *409529_S*, *402364_N*, *400109_S*, we also observed that 10 out of 26 devices have their *Central* method outperforming *NeighborFL L1* over 50% of the times. This is particularly evident for devices such as *402365_S*, *400394_S*, *400045_N*, and *400965_N*. In fact, the *Central* method produces the best results for these four devices compared to *NaiveFL*, *r-NaiveFL*, and *NeighborFL*. Our preliminary hypothesis is that *402365_S*, *400394_S*, and *400045_N* are all located on the same small segment of highway. Their location in the network, as shown in Fig. 5, is a critical junction connecting two major highways, likely experiencing severe bottleneck congestion during rush periods. The prediction plots for *402365_S* and *400394_S* in Fig. 6 show a severe drop in speed around the round 235 period, which is inconsistent with most other sensors. In the future, it would be desirable for the enhanced *NeighborFL* to autonomously recognize such scenarios and avoid performing any type of federation to at least match the performance of the *Central* method. Potential solutions for addressing this issue will be discussed in Section 5.

Table. 5 presents the Average Device MSE values for the 26 devices across the 250 rounds, categorized by the prediction methods. The calculation of those average MSE values follows the same procedure as the values shown in Table. 3. It is worth noting that the *Central* method generally exhibits much larger average MSE values compared to the FL methods. Additionally, we observe that, under both pretrained and non-pretrained settings, the average MSE values for all four configurations of *NeighborFL* (highlighted in green) outperform the baseline methods. Since the average MSE values among the four *NeighborFL* configurations are very close, and different removal methods may be developed to adapt to specific traffic situations, we have omitted the comparison analysis among the four removal methods of *NeighborFL*.

Table 5. Average Device MSE of entire 250 rounds

| Methods | Pretrain MSE | Non-Pretrain MSE |
|---|---:|---:|
| Central | 10.82 | 22.19 |
| NaiveFL | 7.73 | 18.36 |
| r-NaiveFL | 7.44 | 17.7 |
| NeighborFL L1 | 6.77 | 14.99 |
| NeighborFL L3 | 7.11 | 15.8 |
| NeighborFL R1 | 6.57 | 15.43 |
| NeighborFL R3 | 6.98 | 16.11 |

## 5 Discussion and Future Work

In this section, we explore potential solutions for *NeighborFL* to outperform the *Central* method in certain situations and enhance the overall performance of *NeighborFL* as a whole. Additionally, we present two potential future research directions to extend *NeighborFL*.

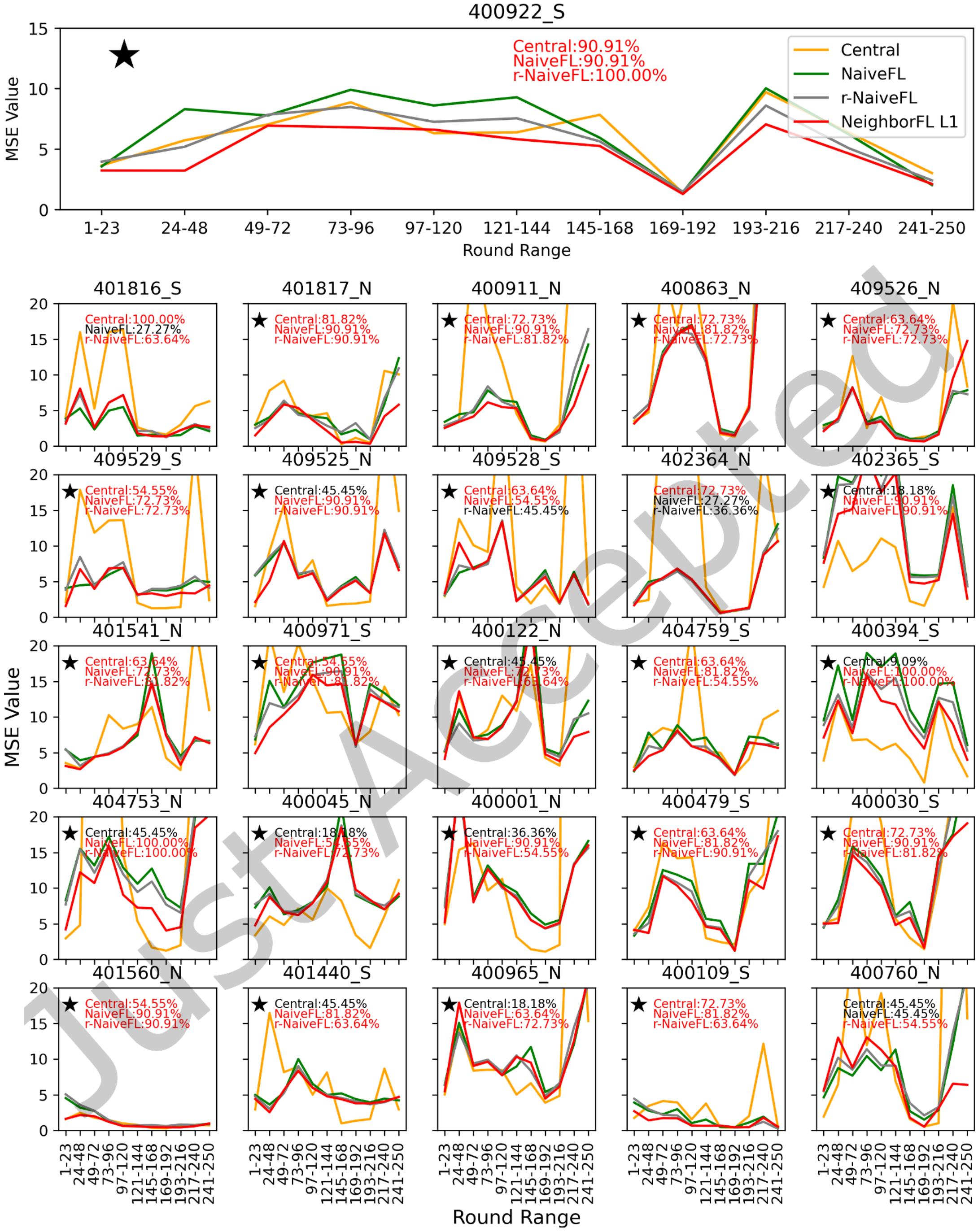

Fig. 7. Smoothed MSE curves of real-time predictions of 26 devices throughout the entire simulation

### 5.1 Improving the performance of NeighborFL

As shown in Fig. 7, there are 10 devices where the *Central* method outperforms all three types of federated methods most of the time, including our best run of *NeighborFL*. However, it is important to note that *NeighborFL L1* still outperforms *NaiveFL* in 9 out of these 10 devices. Despite the fact that the *Central* method exhibits the largest Average Device MSE throughout the entire execution, as shown in Table. 3, we aim to address this issue and elevate the performance of *NeighborFL* to be on par with *Central* for our future work. One primary drawback of *NeighborFL* is the lack of direct consideration of local traffic trends within the candidate radius $r$ when evaluating or removing candidates, instead relying on one-time error evaluation and time- or reputation-based removal. Ignoring the sensor's purpose and the properties of the observed road segment may degrade the model performance for some clients. One potential solution is to reset a device's favorite neighbor set (*FN*) entirely if its real-time prediction error continuously increases, allowing *NeighborFL* to operate similarly to the *Central* method for an extended period and recalibrate by re-evaluating its candidate favorite neighbors from scratch. Another advanced approach is to incorporate local traffic trend analysis and road segment properties into the evaluation criteria for candidate selection. In conclusion, experimenting with novel neighbor evaluation heuristics and/or favorite neighbor removal mechanisms will be key design choices to enhance the performance of *NeighborFL*.

### 5.2 Predicting the impacts of Non-Recurrent Events

Intelligent Transportation Systems often involve predicting the impacts of non-recurrent events such as bad weather conditions or sports events on traffic congestion and delays [2]. Fig. 6 demonstrates that *NeighborFL* exhibits better adaptability to sudden traffic dynamic changes compared to the selected baseline methods. Moreover, with its support for multi-horizon prediction capabilities, we believe that *NeighborFL* has the potential to predict the impacts of non-recurrent events. By training *NeighborFL* on traffic datasets that include non-recurrent events, the system may rapidly adjust a device's *FN* to include only those devices within its radius that would be affected by these changes. This adaptation can lead to more accurate short-term traffic predictions precisely when non-recurrent events occur. Thus, *NeighborFL* shows promise in predicting the effects of such events and improving real-time traffic management.

### 5.3 Detecting malfunctioning detectors

As *NeighborFL* enables devices to remove certain favorite neighbors when real-time prediction errors increase, it opens up the potential for detecting malfunctioning detectors. In practice, a group of devices within a specific area could share information about their removed favorite neighbors. If a particular device consistently appears in most removal reports, there is a high probability that it is a malfunctioning detector. In such cases, the devices can report this finding to the traffic management center, prompting a repair order. Moreover, nearby devices can also be proactive by temporarily excluding the malfunctioning detector from their favorite neighbors set for an extended period. By incorporating such mechanisms, *NeighborFL* contributes to improved traffic management and enhanced overall system reliability.

## 6 Conclusion

The paper presents *NeighborFL*, a decentralized Federated Learning-based approach for real-time traffic prediction, leveraging spatial-temporal relationships among detectors. The methodology enables devices to dynamically allocate personalized groups for model aggregation based on real-time prediction errors, mitigating the non-IID issues of traffic data collected in different locations and enhancing prediction performance by up to 16.9% compared to the naive FL setting. Future research will explore additional neighbor evaluation heuristics and favorite neighbor removal mechanisms to further enhance prediction accuracy, predict impacts of non-recurrent events, and detect malfunctioning detectors. *NeighborFL* shows potential for efficient and adaptable Intelligent Transportation Systems, benefiting commuters and traffic management agencies.

| Symbol | Definition |
|---|---|
| $d_i \in \mathcal{D}$ | The $i$-th traffic device in the Device Set |
| $R_j \in \mathcal{R}$ | The $j$-th communication round |
| $cfn \in d_i.CFN$ | A candidate favorite neighbor $cfn$ belonging to the *Candidate Favorite Neighbors* set of $d_i$. *CFN* is an ordered hash map of {**Key**: a $cfn$'s Device ID, **Value**: the Haversine distance between the $cfn$ and $d_i$}, sorted by value from low to high (close to far Haversine distance). |
| $fn \in d_i.FN$ | A favorite neighbor $fn$ belonging to the *Favorite Neighbors* set (or stack data structure) of $d_i$ |
| $r_i$ | The radius value taken by $d_i$ to form its *CFN* |
| $d^j_{i,eval}$ | A selected $cfn$ by $d_i$ in $R_j$ for evaluation in $R_{j+1}$ |
| $data^j_i$ | The dataset of traffic data points collected by $d_i$ until the end of $R_j$, bounded by *MaxDataSize* |
| *MaxDataSize* | The maximum size of data points stored in $data$ |
| $\tau$ | The number of data points a device collects within a communication round, and also the number of predictions a device makes in $R_2$ and onwards |
| $x$ | A traffic data point. $x^j_{i,m}$ denotes the $m$-th data point collected by $d_i$ in $R_j$ |
| $\mathcal{I}$ | The number of input units in the model used for federated learning, where $\mathcal{I} > 1$ |
| $\mathcal{O}$ | The number of output units in the model used for federated learning, where $\mathcal{O} \geq 1$ |
| $I = \{\mathbf{X}, y\}$ | A data instance in $data$. $\mathbf{X}$ is a vector of data points with size $\mathcal{I}$, and $y$ can be a scalar value or a vector of data points depending on the value of $\mathcal{O}$ |
| $\tilde{\mathbf{X}}$ | The latest $\mathcal{I}$ data points from $data$ a device uses to perform a prediction for the next $\mathcal{O}$ steps |
| $\hat{y}$ | One prediction instance with a size of $\mathcal{O}$ using $\tilde{\mathbf{X}}$ as the input to the model |
| $L$ | A local model. In $R_j$, $L^j_i$ denotes the local model of $d_i$, and $L^j_{i,eval}$ denotes $d^j_{i,eval}$'s local model |
| $G$ | Shared global model produced at the end of a communication round in conventional FL |
| $A$ | An aggregated model. $A^j_i$ denotes the model produced by taking $L^j_i$ and local models from devices in $d_i.FN$, and $A^j_{i,eval}$ denotes the model produced by aforementioned models plus $L^j_{i,eval}$. $A^0_i$ denotes the initial model of $d_i$ before participating in *NeighborFL* |
| $\mathcal{E}$ | The number of local training epochs |
| $P$ | Prediction instances, i.e., a list of $\hat{y}$'s. $P^j_{i,eval}$ denotes the predictions produced by $A^j_{i,eval}$, and $P^j_i$ denotes the predictions produced by $A^j_i$ or $G$ |
| $y$ | One ground truth data instance with a size of $\mathcal{O}$ from the real-time updated $data$ |
| $Y$ | Ground truth instances, i.e., a list of $y$'s. $Y^j_i$ represents the data values collected by $d_i$ in $R_j$ |
| $E$ | Prediction error, which could be MAE, MRE, MSE, RMSE, etc. $E^j_{i,eval}$ denotes the prediction error calculated by $P^j_{i,eval}$ and $Y^j_i$, and $E^j_i$ is the error calculated by $P^j_i$ and $Y^j_i$ |
| $\sim fn$ | A removed favorite neighbor from a device's *FN* set |
| $\nu$ | The number of continuous rounds of a device's real-time prediction error increase required to trigger a favorite neighbor removal action |
| $d_i$.rep_book | A hash map for $d_i$ to keep track of the reputation of its $cfn$'s. The reputation of a $cfn$ is calculated as $E_i - E_{i,eval} = \Delta E$ when it is being evaluated as a $d_{i,eval}$ |
| $d_i$.last_try_round | A hash map for $d_i$ to keep track of the (re-)evaluation round for $d_{i,eval}$ when $d_{i,eval}$ is being removed or skipped from being added to $d_i.FN$ due to failure in reducing the prediction error |
| $d_i$.retry_interval | A hash map for $d_i$ to keep track of and examine the retry interval for a $cfn$ when $cfn$ is being considered for selection as a $d_{i,eval}$ |

Table 6. List of Key Mathematical Notations